\documentclass{article}

    \usepackage[final]{neurips_2022}

\usepackage[utf8]{inputenc} 
\usepackage[T1]{fontenc}    
\usepackage{url}            
\usepackage{booktabs}       
\usepackage{amsfonts}       
\usepackage{nicefrac}       
\usepackage{microtype}      

\usepackage{dsfont,amsmath,amsfonts,bm,amssymb,mathtools,amsthm}

\usepackage{color}
\usepackage{colortbl}
\usepackage[table]{xcolor}
\usepackage{epsfig}
\usepackage[outdir=./]{epstopdf}

\usepackage{microtype}
\usepackage{lipsum}
\usepackage{subcaption}
\usepackage{latexsym}
\usepackage{sidecap}
\usepackage{caption}
\usepackage{booktabs} %
\usepackage{nicefrac}       %
\usepackage{comment}
\usepackage{enumitem}
\usepackage{wrapfig,tikz}
\usepackage{longtable,fancyhdr}
\usepackage{adjustbox, bigstrut, tabularx, multirow, makecell, diagbox}
\usepackage{graphicx,float}
\usepackage{stackrel}
\usepackage{theoremref}
\usepackage{cases}
\usepackage{stmaryrd}
\usepackage{mathabx}
\usepackage{dsfont}
\usepackage{arydshln}

\usepackage{url} 

\usepackage{nicematrix}

\usepackage{tikz}
\usetikzlibrary{positioning}

\usepackage[toc,page,header]{appendix}
\usepackage{minitoc}

\usepackage[pagebackref,breaklinks,colorlinks]{hyperref}

\usepackage{algorithm}
\usepackage{algpseudocode}

\definecolor{red_}{RGB}{214, 39, 40}
\definecolor{green_}{RGB}{214, 39, 40}
\definecolor{blue_}{RGB}{158, 218, 229}
\definecolor{orange_}{RGB}{255, 187, 120}
\definecolor{mutedlightgreen}{RGB}{219, 245, 206}

\usepackage[capitalize]{cleveref}
\crefname{section}{Sec.}{Secs.}
\Crefname{section}{Section}{Sections}
\Crefname{table}{Table}{Tables}
\crefname{table}{Tab.}{Tabs.}
\crefname{algorithm}{Alg.}{Algs.}

\def\thanks#1{\protected@xdef\@thanks{\@thanks
\protect\footnotetext{#1}}}
\makeatother

\newtheorem{theorem}{Theorem}

\makeatletter
\usepackage{xspace} 
\def\@onedot{\ifx\@let@token.\else.\null\fi\xspace}
\DeclareRobustCommand\onedot{\futurelet\@let@token\@onedot}

\def\eg{\emph{e.g}\onedot}

\def\ie{\emph{i.e}\onedot}

\begin{document}

\title{Generator Born from Classifier}

\author{\bf Runpeng Yu \quad
Xinchao Wang$^{\dagger}$ 
\\
National University of Singapore\\
{\tt\small r.yu@u.nus.edu} \quad {\tt\small xinchao@nus.edu.sg}
}

\maketitle

\begin{abstract}
    In this paper, we make a bold attempt toward an ambitious task:
    given a pre-trained classifier, we aim to reconstruct an image generator, without relying on any data samples.
    From a black-box perspective, this challenge seems intractable, since 
    it inevitably involves  identifying the inverse function for a classifier,
    which is, by nature, an information extraction process.
    As such, we resort to leveraging the 
    knowledge encapsulated within the parameters of the neural network.
    Grounded on the theory of Maximum-Margin Bias of gradient descent, we propose a novel learning paradigm, in which the generator is trained to ensure that the convergence conditions of the network parameters are satisfied over the generated distribution of the samples. Empirical validation from various image generation tasks substantiates the efficacy of our strategy.

\end{abstract}
\section{Introduction}

\let\thefootnote\relax\footnotetext{$^{\dagger}$ Corresponding author.}

The majority of machine learning research bifurcates into two distinct branches of study: the predictive task and the generative task. Given the input $\bm{x}$ and the label $y$, the former one  focuses on the training of a high-performing classifier or regressor, which approximates  $p(y|\bm{x})$ ~\citep{attention,vit,jing2023deep}, whereas the latter one aims to train a generative model capable of sampling from $p(\bm{x}|y)$ or $p(\bm{x},y)$~\citep{gan}. The gap between the predictive and the generative models, as a result, predominantly arises from the lack of information in the predictive models about the marginal distribution $p(\bm{x})$. In the realm of deep neural networks, however, the over-parameterization leads to the overfitting on the training distribution and the memorization of the training samples~\citep{memorization_1,memorization_2,memorization_3}, which, in turn, make the network  implicitly retain information about $p(\bm{x})$. With this component in hand, it prompts the question of whether it is feasible to derive a generative model from a predictive one.

In this paper, we explore this novel task,
which attempts to
learn a generator directly from a pre-trained classifier, \emph{without} the assistance of any training data. 
Unarguably, this is a highly ambitious task
with substantial difficulty, as either explicitly extracting information about $p(\bm{x})$ from a pre-trained classifier or directly solving this inverse problem
from classifier to generator poses significant challenges. 
Despite these challenges, the value of this task lies in its potential to offer a new approach to training generators that mitigates the direct dependence on large volumes of training data. This provides a possible solution for learning tasks in scenarios where data is scarce or unavailable. Moreover, this task presents a novel way to utilize and analyze the pre-trained predictive models, facilitating our understanding of the encoded information within the parameters.

To this end, we propose a novel learning scheme.
Our approach is grounded in the theory of Maximum-Margin Bias of gradient descent, which demonstrates that the parameters of a neural network trained via gradient descent will converge to the solution of a specific optimization problem. This optimization problem minimizes the norm of the neural network parameters while maximizing the classification margin on the training dataset. The necessary condition for the solution of this optimization problem constructs a system of equations, describing the relationship between the pre-trained neural network parameters and the training data distribution. 

Since our aim is to learn a generator from the pre-trained classifier parameters,  the generator is therefore expected to approximate a distribution that satisfies the necessary condition of this optimization problem. To accomplish this, we design the loss function for training the generator and the corresponding training algorithm based on the necessary condition. The entire training process does not rely on any training data; all available information related to the pre-trained data is encapsulated within the parameters of the pre-trained classifier. 

The intuition behind our design is twofold: on one hand, the generator should guarantee that the pre-trained classifier performs well under the data distribution it approximated, and on the other hand, the generator should ensure that the current classifier parameters are the convergence point of the gradient descent algorithm under the data distribution it approximated. 
It's noteworthy that the original data distribution naturally satisfies these conditions. Therefore, we anticipate that employing the proposed method will guide the generator to discover the original data distribution.

\begin{figure}
  \centering
  \begin{subfigure}{0.4\linewidth}
    \includegraphics[height=5.3em]{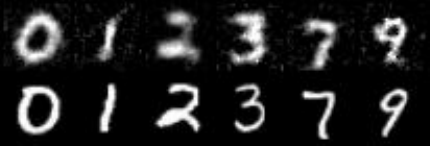}
    \caption{MNIST}
    \label{fig:head:a}
  \end{subfigure}
  \begin{subfigure}{0.53\linewidth}
    \includegraphics[height=5.3em]{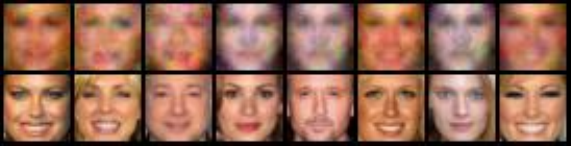}
    \caption{CelebA}
    \label{fig:head:b}
  \end{subfigure}
  \caption{Images produced by the generator trained only using pre-trained classifier. The generated images, positioned in the first row, are accompanied by their nearest neighbors from the original dataset, displayed in the second row.}
  \label{fig:head}\vspace{-2em}
\end{figure}

We conduct experiments on commonly used image datasets. \cref{fig:head} shows some generated images of the MNIST and CelebA datasets. Remarkably, even trained without access to the original data, the generator is able to perform conditional sampling and generate the digits and faces.

Our contribution is therefore a novel approach that, for the first time, attempts to train a generator from a pre-trained classifier without utilizing training data. The proposed approach produces encouraging results on synthetic and real-world images. 

A list of symbols utilized in this paper and corresponding descriptions can be found in the Appendix.

\section{Related Work}
\textbf{Generative Adversarial Networks.}
Generative Adversarial Network (GAN) consists of a generator and a discriminator collectively optimizing a minimax problem to learn and replicate the original data distribution ~\citep{gan}. Numerous extensions of the original GAN have been investigated, including functionality enhancement~\citep{infogan,cgan,acgan,bigan}; architecture optimization and scaling~\citep{progan,biggan,lagan};
and training loss design~\citep{fgan,wgan,wgangp}.
Owing to their superior generation quality, GANs have found wide-ranging applications in image synthesis~\citep{gan_imagegenerate_1}, blending~\citep{gan_blending_1}, inpainting~\citep{gan_inpainting,gan_inpainting_2,gan_inpainting_3}, super-resolution~\citep{srgan}, denoising~\citep{gan_denoising-1}; image-to-image translation~\citep{cyclegan,pix2pix}; 3D object generation~\citep{gan_3d_1}; video generation~\citep{gan_video_1,gan_video_2}; etc.

Both our work and GAN require an additional classifier to guide the training of the generation and provide a measure of authenticity for generated data. However, in the GAN framework, the classifier is trained concurrently with the generator, with the explicit goal of discerning the quality of generated results. In contrast, our method utilizes a pre-trained classifier, which can be arbitrary, and its training objective is to maximize classification accuracy, not to judge the quality of generated results. This imbues our approach with considerable flexibility. Furthermore, during the training of GAN, the generator has access to training data. However, in our task, the training data is not available to the generator, which makes our task harder.

\textbf{Feature Visualization and Model Inversion.} Besides our task, neural network feature visualization and model inversion share the objective of extracting information relevant to the training data from pre-trained classifiers. Neural network feature visualization is a technique aimed at identifying input data that maximally activates specific neurons or layers within the network, thereby providing insights into the patterns or features that the network is primed to recognize.~\citep{fv_1,fv_2,fv_3} Model inversion in neural networks refers to the process of inferring or reconstructing input data given the trained model.~\citep{mi_1} In the realm of adversarial attacks, model inversion is deployed to discover sensitive information about the training data from the model's outputs.~\citep{mi_attack,mi_attack_1,mi_attack_2}

Unlike these tasks, where an independent gradient optimization is required for each generation, our goal is to develop a generator capable of sampling from the training data distribution. While there is research in neural network feature visualization and model inversion that utilizes a generative model to assist in the restoration of specific training data, these works typically employ the generator more as a prior for the recovery process.~\citep{fv_gan_1,mi_gan_1,mi_gan_2,mi_gan_3} The training of such a generator necessitates additional training data, which should be similar to or encompass the original training data used for the classifier. In contrast, our approach directly derives a generator from the classifier without the utilization of extraneous training data.

\textbf{Energy-Based Model.} Within the framework of energy-based models, pre-trained classifiers have also been demonstrated to be capable of acting as a conditional probability distribution, generating data samples. ~\citep{classifier_energy_based_model,generative_classifier_joint_train_1,generative_classifier_joint_train_2} 
Although the energy-based models theoretically bridge predictive and generative models, practical implementation for sample generation based on it still relies on an optimization target and executes a multi-step gradient optimization process, akin to model inversion. In contrast, our objective is to develop a generator with the ability for random or condition-based sampling, procuring data samples directly through the forward pass of the neural network.

\textbf{Maximum-Margin Bias of Gradient Descent.} Our work is based on the study of the Maximum-Margin Bias of gradient descent. These investigations primarily seek to elucidate why gradient descent algorithms are capable of learning models with robust generalization capabilities, even in the absence of additional regularization constraints. It has been discovered that under the guidance of gradient descent, the parameters of a neural network converge to a solution to an optimization problem aiming to maximize the classification boundary while concurrently minimizing the norm of the network parameters. Initial studies focused on linear logistic regression models~\citep{mm_logist_1,mm_logist_2}, which were subsequently extended to homogeneous neural networks~\citep{mm_homo_1,mm_homo_2,mm_homo_3,mm_homo_4,mm_homo_5,mm_homo_6} and, more recently, to a broader class of quasi-homogeneous neural networks~\citep{mmb}. The theory of Maximum-Margin Bias has also been leveraged by \citet{DataRec} to recover training data. However, their objective was the restoration of data rather than training a generator, and their work was exclusively confined to binary classification datasets and fully connected networks without bias terms.
\section{Preliminary}

To extract the information about the training dataset from the parameters of the pre-trained classification model, we leverage the theory of Maximum-Margin Bias. 

Let $\Phi(\cdot;\zeta):\mathcal{R}^d \to \mathcal{Y}$ denote the classifier parameterized by $\zeta$ and trained on multi-class classification dataset $D = \{(\bm{x}_i,y_i)\}_{i=1}^N$ with each $(\bm{x}_i,y_i)\in\mathcal{R}^d \times \mathcal{Y}$. let $L(\zeta) := \sum_{i=1}^N l( \Phi(\bm{x}_i;\zeta), y_i)$ denote the standard cross-entropy loss of $\Phi$ on $D$. 
To extend the applicability of Maximum-Margin Bias analysis to various neural network structures, the definition of the \textit{$\Lambda$-quasi-homogeneous model} is introduced. For a (non-zero) positive semi-definite diagonal matrix $\Lambda$
, a model $\Phi(\cdot;\zeta)$ is $\Lambda$-quasi-homogeneous if the model output scales as $\Phi(\bm{x};\psi_\alpha(\zeta)) = e^{\alpha}\Phi(\bm{x};\zeta)$ for all $\alpha \in \mathcal{R}$ and input $\bm{x}$, when the parameter scales as $\psi_\alpha(\zeta) := e^{\alpha \Lambda}\zeta$. Many commonly used neural network architectures satisfy the definition of $\Lambda$-quasi-homogeneous model, such as  convolutional neural networks and fully connected networks with biases, residual connections, and normalization layers. The seminorm of $\zeta$ corresponding to $\Lambda$ is defined as 
$||\zeta||^2_{\Lambda}:=\zeta^T \Lambda \zeta$. Let $\lambda_{max}:=\max_{j} \Lambda_{jj}$ denote the maximal element in $\Lambda$, $\tilde{\Lambda}$ is the matrix setting all elements less than $\lambda_{max}$ to $0$, $\ie$, $\tilde{\Lambda}_{jj} = \lambda_{max}$ if $\Lambda_{jj} = \lambda_{max}$, otherwise, $\tilde{\Lambda}_{jj} = 0$. Accordingly, the seminorm of $\zeta$ corresponding to $\tilde{\Lambda}$ is defined as 
$||\zeta||^2_{\tilde{\Lambda}}:=\zeta^T \tilde{\Lambda} \zeta$.
The normalized parameters is defined as $\bar{\zeta}:=\psi_{\tau}(\zeta)$, such that $\|\bar{\zeta}\|_\Lambda^2 = 1$.
The Quasi-Homogeneous Maximum-Margin Theorem states as follows.
\begin{theorem}[Paraphrased from \citep{mmb}]
\label{thm:mmb}
Let $\Phi(\cdot;\zeta)$ denote a $\Lambda$-quasi-homogeneous classifier trained on $D$ with cross-entropy loss $L$. Assume that: (1) for any fixed $\bm{x}$, $\Phi(\bm{x};\zeta)$ is locally Lipschitz and admits a chain rule~\citep{DavisDKL20,LyuL20}; (2) the learning dynamic is described by a gradient flow~\citep{LyuL20}; (3) $\lim_{t \to \infty}\bar{\zeta}(t)$ exists; (4) $\exists\kappa > 0$ such that only $\zeta$ with $||\zeta||_{\Lambda_{\max}} \ge \kappa$ separates the training data; and (5) $\exists t_0$ such that $L(\zeta(t_0))<N^{-1}\log 2$.
$\exists \alpha\in\mathcal{R}$ such that $\Tilde{\zeta}:=\psi_\alpha(\lim_{t \to \infty}\bar{\zeta}(t))$ is a first-order stationary point of the following maximum-margin problem
\begin{subequations}\label{eq:opt}
    \begin{align}
        \underset{\zeta'}{\text{min}}
         &\qquad \frac{1}{2} ||\zeta'||^2_{\tilde{\Lambda}}
        \\
        \text{s. t.} &\qquad  \min_{c\in\mathcal{Y}/\{y_i\}} \Phi_{y_i}(x_i;\zeta')-\Phi_{c}(x_i;\zeta') \geq 1\quad\forall i\in[N],
    \end{align}
\end{subequations}

where $\Phi_{c}(\cdot;\zeta')$ is the prediction of $\Phi$ for the class $c\in\mathcal{Y}$.
\end{theorem}

\section{Method}

\cref{thm:mmb} implies that the neural network parameters converge to the first-order stationary point (or the Karush–Kuhn–Tucker point (KKT) point) of the optimization problem in \cref{eq:opt}.
Let $\{\mu_{ic}\}_{i\in[N],c\in\mathcal{Y}/\{y_i\}}$ denote the set of KKT multipliers, the KKT condition can be written as follows.
\begin{subequations}\label{eq:kkt}
\begin{align} 
    &\tilde{\Lambda} \tilde{\zeta} = \sum_{i\in[N]}\sum_{c\in\mathcal{Y}/\{y_i\}}\mu_{ic}[\nabla_{\tilde{\zeta}}\Phi_{c}(x_i;\tilde{\zeta})-\nabla_{\tilde{\zeta}}\Phi_{y_i}(x_i;\tilde{\zeta})];
    \label{eq:stationary}\\[0.5em]
    & \text{for all } i \in [N], \text{and } c\in\mathcal{Y}/\{y_i\}:\nonumber\\[0.3em]
    & \qquad \Phi_{y_i}(x_i;\tilde{\zeta})-\Phi_{c}(x_i;\tilde{\zeta}) \geq 1, 
    \label{eq:primfeas} \\[0.1em]
    & \qquad \mu_{ic} \geq 0, 
    \label{eq:dualfeas}\\[0.1em]
    & \qquad \mu_{ic}[1+\Phi_{c}(x_i;\tilde{\zeta})-\Phi_{y_i}(x_i;\tilde{\zeta})]= 0, 
    \label{eq:compslack}
\end{align}
\end{subequations}
where the \cref{eq:stationary,eq:primfeas,eq:dualfeas,eq:compslack} are known as the stationarity condition, primal feasibility condition, dual feasibility condition, and the complementary slackness condition, respectively.

The intuition behind the stationarity condition can be summarized as follows. 
According to the value of the corresponding element in $\Lambda$, the neural network parameters can be divided into two groups: the first group $\mathrm{Z}_1 := \{\tilde{\zeta}_j|\Lambda_{jj} = \lambda_{max}\}$ includes all the parameters whose corresponding elements in $\Lambda$ are equal to $\lambda_{max}$, and the second set $\mathrm{Z}_2 := \{\tilde{\zeta}_j|\Lambda_{jj} \neq \lambda_{max}\}$ contains all the parameters whose corresponding elements in $\Lambda$ are not equal to $\lambda_{max}$. 
The stationarity condition in \cref{eq:stationary} evaluates 
the linear combination of the derivatives of 
the neural network output 
corresponding to the parameters, 
with KKT multipliers $\{\mu_{ic}\}_{i\in[N],c\in\mathcal{Y}/\{y_i\}}$ as coefficients of the combination. 
Such a linear combination of the derivatives corresponding to the parameters in $\mathrm{Z}_1$ is equal to the parameters in $\mathrm{Z}_1$.
In contrast, such a linear combination of the derivatives corresponding to the parameters in $\mathrm{Z}_2$ is equal to a zero vector. 

The intuition behind other conditions can be summarized as follows.
According to the complementary slackness condition in \cref{eq:compslack}, $\mu_{ic}$ is nonzero only when $\Phi_{y_i}(x_i;\tilde{\zeta})-\Phi_{c}(x_i;\tilde{\zeta})=1$. Two conditions are required for $\mu_{ic}$ to be nonzero. First, for a pair of $(i,c)$, $\mu_{ic}$ can only be nonzero when $c$ is the class with the second largest predicted probability for the sample $x_i$, \ie, $c \in \mathcal{S}_i$, where  $\mathcal{S}_i := \{ y|\Phi_{y}(x_i;\tilde{\zeta})=\max_{y'\in\mathcal{Y}/\{y_i\}}\Phi_{y'}(x_i;\tilde{\zeta})\}$. Second, according to primal feasibility in \cref{eq:primfeas}, the minimum possible value of $\Phi_{y_i}(x_i;\tilde{\zeta})-\Phi_{c}(x_i;\tilde{\zeta})$ is $1$. Therefore, for a pair of $(i,c)$, $\mu_{ic}$ can only be nonzero when the margin between the true class $y_i$ and the class with the second largest predicted probability for the sample $x_i$ is minimum. 

The evaluation of $\nabla_{\tilde{\zeta}}\Phi(\cdot;\tilde{\zeta})$ and $\Phi(\cdot;\tilde{\zeta})$ requires to first scale all parameters of a pre-trained network. For the convenience of practical implementation and following derivation, we transform $\tilde{\zeta}$ back to $\zeta$ using the definition of the quasi-homogeneous function. The KKT conditions in \cref{eq:kkt} are rewritten as:
\begin{subequations}\label{eq:kkt:zeta}
\begin{align} 
    &\bar{\Lambda}\zeta = \sum_{i\in[N]}\sum_{c\in\mathcal{Y}/\{y_i\}}\mu_{ic}[\nabla_{\zeta}\Phi_{c}(x_i;\zeta)-\nabla_{\zeta}\Phi_{y_i}(x_i;\zeta)]; 
    \label{eq:stationary:zeta}\\[0.5em]
    & \text{for all } i \in [N], \text{and } c\in\mathcal{Y}/\{y_i\}:&\nonumber\\[0.3em]
    & \qquad \Phi_{y_i}(x_i;\zeta)-\Phi_{c}(x_i;\zeta) \geq e^{-\alpha}, 
    \label{eq:primfeas:zeta} \\[0.1em]
    & \qquad \mu_{ic} \geq 0, 
    \label{eq:dualfeas:zeta}\\[0.1em]
    & \qquad \mu_{ic}[e^{-\alpha}+\Phi_{c}(x_i;\zeta)-\Phi_{y_i}(x_i;\zeta)]= 0, 
    \label{eq:compslack:zeta}
\end{align}
\end{subequations}
where the new scaling parameter $\bar{\Lambda}:=\tilde{\Lambda} e^{\alpha(2\Lambda-\mathbf{I})}$.

\begin{wrapfigure}{r}{0.5\textwidth}
    \centering
  \vspace{-1em}
	\includegraphics[width=\linewidth]{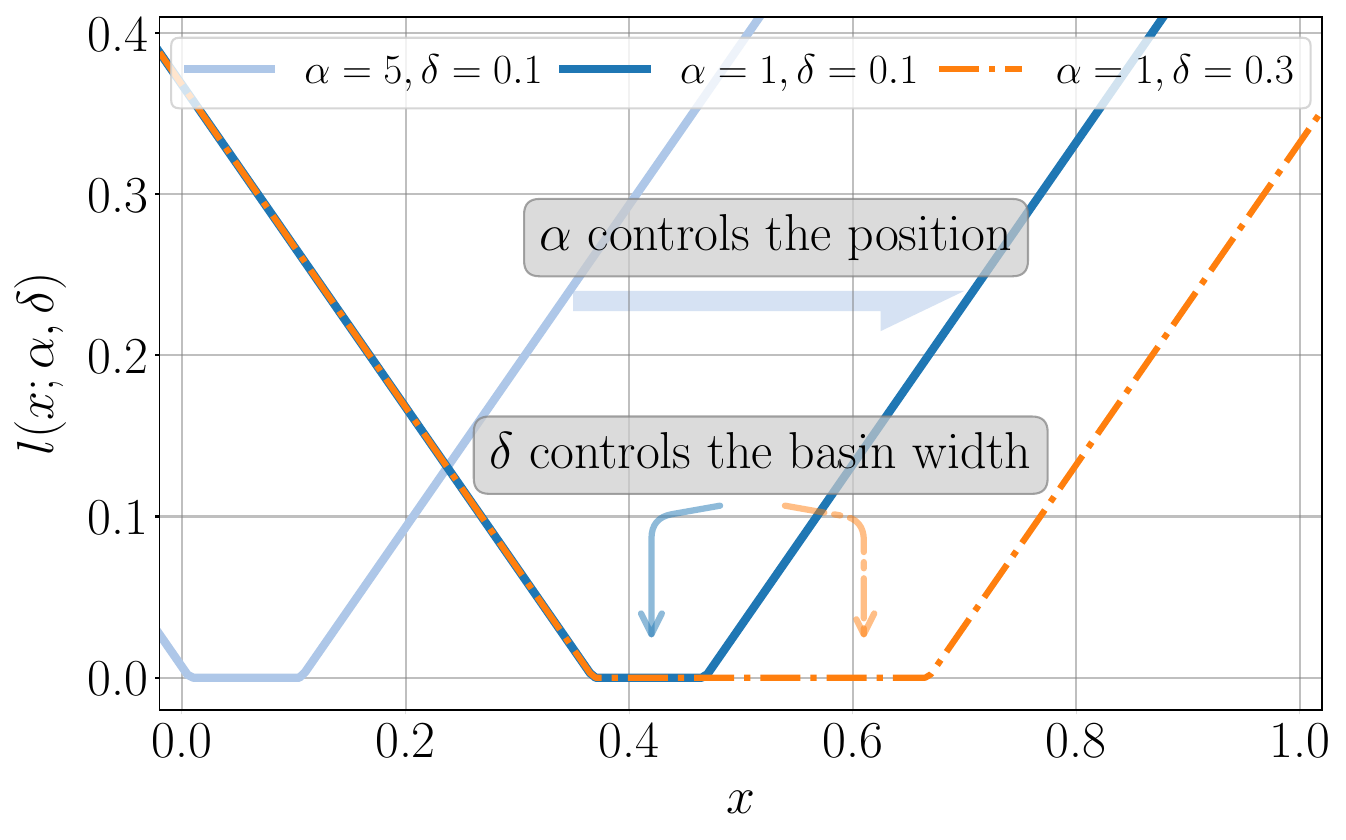}
  \vspace{-2em}
	\caption{The U-shaped duality loss $l(x;\alpha,\delta):=\max\left(x\!-\!e^{-\alpha}\!-\!\delta,0\right)\!-\!\min\left(x\!-\!e^{-\alpha},0\right)$.}\label{fig:Lduality}
 \vspace{-1em}
\end{wrapfigure}

\textbf{From KKT condition to loss function.}   Given only the pre-trained neural network $\Phi(\cdot;\zeta)$, the undetermined parts in the KKT conditions in \cref{eq:kkt:zeta} include the KKT multipliers, a set of $(x,y)$ pairs,
and constant $\alpha$. 
Regarding the KKT multiplier a predictable objective, We use a neural network to learn it. Given an approximate distribution of the discrete random variable $y$, we sample $y$ directly. Our goal is to train a conditional generator $g$ parameterized by $\theta$ to generate input $x=g(\epsilon,y;\theta)$ given the corresponding label $y$ and random noise $\epsilon$. We also treat $\alpha$ as a learnable parameter, which will be discussed in detail later. In the following paragraphs, we first discuss how to design the loss function for learning these parameters and training the generator.

First, we discuss how to ensure that the generated samples satisfy the stationary condition in \cref{eq:stationary:zeta}. The evaluation of the right-hand side of \cref{eq:stationary:zeta} requires generating a  dataset with a fixed number of samples every time to calculate the summation. Alternatively, we divide both sides of \cref{eq:stationary:zeta} by the number of training samples $N$, which converts the sum on the right-hand side to an expectation:
\begin{align}
    &\frac{1}{N}\bar{\Lambda}\zeta = \mathbb{E}_{x,y}\big[\sum_{c\in\mathcal{Y}/\{y\}}\mu_{c}[\nabla_{\zeta}\Phi_{c}(x;\zeta)-\nabla_{\zeta}\Phi_{y}(x;\zeta)]\big].
    \label{eq:stationary:rescale}
\end{align}
Thus, according to the law of large numbers, the expectation can be estimated by the empirical average over a batch of $M$ random samples, where $M$ is a hyper-parameter. To ensure that the rescaled stationary condition in \cref{eq:stationary:rescale} can be satisfied, we use the following $L_{stationarity}$ to minimize the norm of the difference between both sides of \cref{eq:stationary:rescale}.
\begin{equation}
    L_{stationarity}(\theta,\eta) := ||\frac{1}{N}\bar{\Lambda}\zeta - \frac{1}{M}\sum_{i\in[M]}\sum_{c\in\mathcal{Y}/\{y_i\}}\mu_{ic}[\nabla_{\zeta}\Phi_{y_i}(x_i;\zeta)-\nabla_{\zeta}\Phi_{c}(x_i;\zeta)]||.
\end{equation}

Conditions in \cref{eq:primfeas:zeta,eq:dualfeas:zeta,eq:compslack:zeta} constrain the valid values of the KKT multipliers. Accordingly, in order to satisfy the positivity condition of the KKT multipliers in dual feasibility, instead of directly optimizing the KKT multipliers, we use the proxy variables $\mu_{ic}'$, define $\mu_{ic} := ReLU(\mu_{ic}')$. For each generated sample $x_i$, we set up a $\mu_{i}'\in\mathcal{R}^{|\mathcal{Y}|}$ with $\mu_{ic}'$ is its $c$-th element. We learn $\mu_{i}'=h(x_i,y_i;\eta)$ by network $h$ parameterized by $\eta$. To approximate primal feasibility and complementary slackness, we require each generated sample $x_i$ to satisfy $0\leq \Phi_{y_i}(x_i;\zeta) - \Phi_{c}(x_i;\zeta) - e^{-\alpha} \leq \delta$, where $0<\delta\ll1$ is a hyper-parameter to ensure numerical stability. We minimize $L_{duality}$ to approximate this constraint. \cref{fig:Lduality} illustrates the shape of $L_{duality}$.
\begin{align}
    L_{duality}(\theta,\alpha):= \frac{1}{M}\sum_{i\in[M]}\sum_{c\in\mathcal{S}_i} &[\max\left(\Phi_{y_i}(x_i;\zeta) - \Phi_{c}(x_i;\zeta) - e^{-\alpha}-\delta,0\right)\nonumber\\
    &\ \ -\min\left(\Phi_{y_i}(x_i;\zeta) - \Phi_{c}(x_i;\zeta) - e^{-\alpha},0\right)]
\end{align}

\begin{wrapfigure}{r}{0.48\textwidth}
    \centering
	\includegraphics[width=\linewidth]{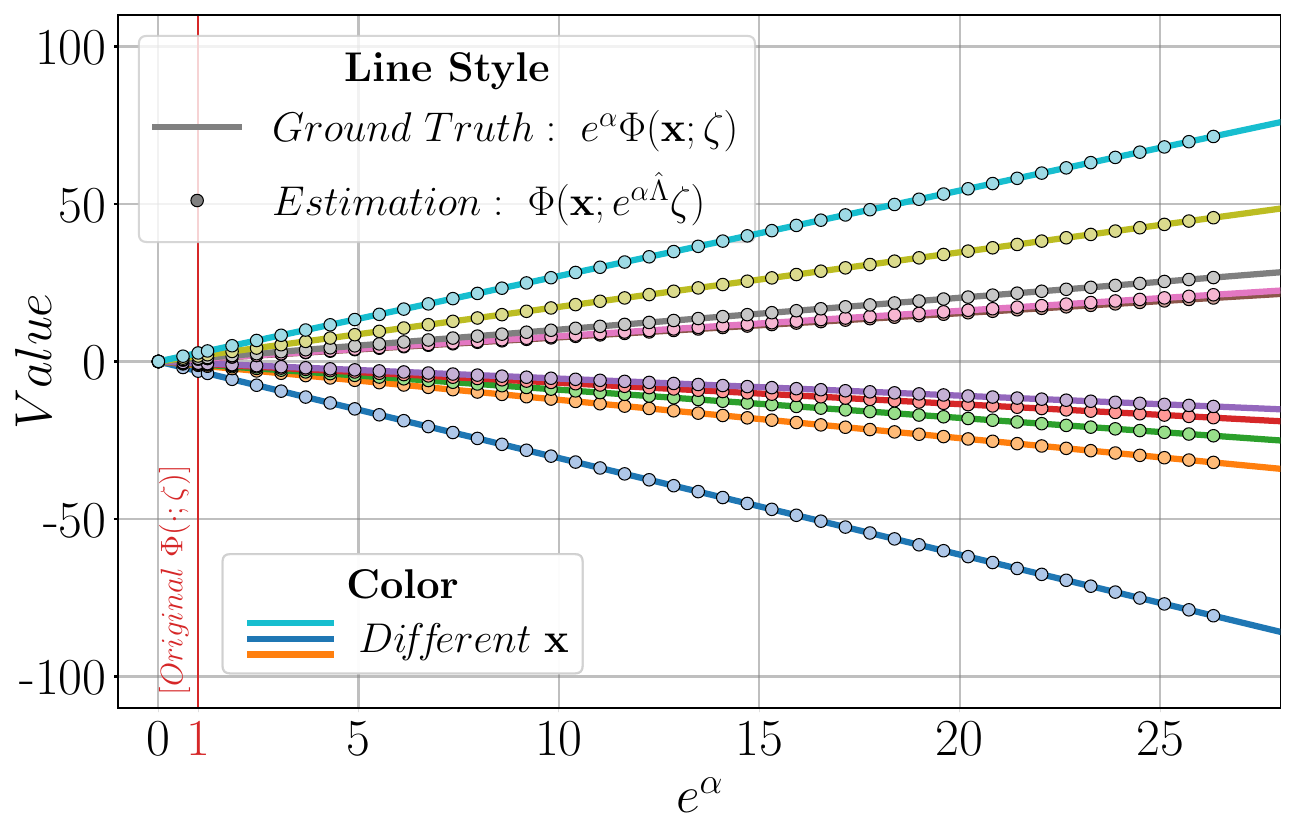}
  \vspace{-2em}
	\caption{The plot of $\Phi(\bm{x};e^{\alpha\hat{\Lambda}}\zeta)$ and $e^{\alpha}\Phi(\bm{x};\zeta)$, with $\hat{\Lambda}$ as the estimation of $\Lambda$. Color distinction signifies diverse input $\bm{x}$. The coincidence of solid and dotted lines validates the precise estimation of $\Lambda$, because, by definition, an accurate estimation  fulfills $\Phi(\bm{x};e^{\alpha\hat{\Lambda}}\zeta)=e^{\alpha}\Phi(\bm{x};\zeta)$.}\label{fig:Lambda_poc}
\end{wrapfigure}

Final loss is the combination of $L_{lagrange}$ and $L_{duality}$  balanced by hyper-parameter $\beta$:
\begin{equation}
    L = L_{stationarity}(\theta,\eta) + \beta L_{duality}(\theta,\alpha).
\end{equation}

\textbf{Determine $\Lambda$. } 
To compute $L_{stationarity}$, it is necessary to determine the $\Lambda$ of the quasi-homogeneous function $\Phi(\cdot;\zeta)$. 
Here, we will introduce two properties of quasi-homogeneous functions and demonstrate how to construct a system of linear equations based on these properties to efficiently solve for $\Lambda$. 
First, taking the derivative of $\Phi(\bm{x};\psi_\alpha(\zeta))$ corresponding to $\zeta$, we have 
\begin{equation}
\nabla_{\psi_\alpha(\zeta)}\Phi(\bm{x};\psi_\alpha(\zeta))=e^{\alpha(\mathbf{I}-\Lambda)}\nabla_{\zeta}\Phi(\bm{x};\zeta). \label{eq:derivative_zeta}
\end{equation}
This indicates that, for any parameter $\zeta_i$ whose corresponding $\Lambda_{ii}\neq1$, the derivative of $\Phi(\bm{x};\zeta)$ corresponding to $\zeta_i$, denoted by $\nabla_{\zeta_i}\Phi(\bm{x};\zeta)$, is a $\Lambda'$-quasi-homogeneous function with $\Lambda' = \frac{1}{1-\Lambda_{ii}}\Lambda$. Second, taking the derivative $\nabla_{\alpha}\Phi(\bm{x};\psi_\alpha(\zeta))$ at $\alpha=0$, we have:
\begin{align}
&\zeta^T\Lambda\nabla_{\zeta}\Phi(\bm{x};\zeta) = \Phi(\bm{x};\zeta).~&\text{(derivative equation of $\Phi$ at $\bm{x}$)}& \label{eq:cal_Lambda}
\end{align}
For clear notation, we refer to the above \cref{eq:cal_Lambda} as the derivative equation of $\Phi$ at $\bm{x}$.
which is a linear equation about $\Lambda$ and the coefficients can be calculated conveniently given the pre-trained model.

As these properties of quasi-homogeneous function are independent of input $\bm{x}$, we can establish a system of equations by evaluating the derivative equation of $\Phi$ and its higher order derivatives at a set of random samples $\{x_i\}_{i\in[K]}$, \eg,
\begin{align}\label{eq:SE_Lambda}
\left\{\begin{aligned}
&&\zeta^T\Lambda\nabla_{\zeta}\Phi(\bm{x}_1;\zeta) &=& &\Phi(\bm{x}_1;\zeta),~&\text{(derivative equation of $\Phi$ at $\bm{x}_1$)} \\
&&\zeta^T\Lambda\nabla_{\zeta}\Phi(\bm{x}_2;\zeta) &=& &\Phi(\bm{x}_2;\zeta),~&\text{(derivative equation of $\Phi$ at $\bm{x}_2$)} \\[-0.5em]
&&\vdots \qquad\qquad&&&\quad\ \vdots\quad& \\[-0.5em]
&\frac{1}{1-\Lambda_{11}}\hspace{-10pt}&\zeta^T\Lambda\nabla^2_{\zeta\zeta_1}\Phi(\bm{x}_K;\zeta) &=&  \nabla_{\zeta_1}&\Phi(\bm{x}_K;\zeta),~&\text{(derivative equation of $\nabla_{\zeta_1}\Phi$ at $\bm{x}_K$)} \\[-0.7em]
&&\vdots \qquad\qquad&&&\quad\ \vdots\quad& \\
\end{aligned}\right..
\end{align}
Then, the $\Lambda$ can be calculated from it. As a proof of concept, the estimated results of a two-layer fully-connected network with ReLU activation are shown in \cref{fig:Lambda_poc}. The architecture of the network is 
\begin{tikzpicture}[baseline=-0.25em]
\node[rectangle, rounded corners, draw=black, fill=blue_!50, inner sep=2pt, font=\footnotesize] (linear1) at (0,0) {$Linear(2,10)$};
\node[font=\footnotesize] (arrow1) at (1.25,0) {$\rightarrow$};
\node[rectangle, rounded corners, draw=black, fill=orange_!50, inner sep=2pt, font=\footnotesize] (relu) at (2.1,0) {$ReLU()$};
\node[font=\footnotesize] (arrow2) at (2.96,0) {$\rightarrow$};
\node[rectangle, rounded corners, draw=black, fill=blue_!50, inner sep=2pt, font=\footnotesize] (linear2) at (4.19,0) {$Linear(10,1)$};
\end{tikzpicture}.

\textbf{Determine $\alpha$. } 
To compute $L_{duality}$, it is necessary to determine the $\alpha$. Shown in the proof of \cref{thm:mmb}~\citep{mmb},
the value of $\alpha$ depends on the minimum classification margin $q_{min}$ of the (normalized) neural network on the training dataset, in the case of a multi-class problem:
\begin{equation}
    q_{min} = \min_{i\in[N]}\min_{c\in[C]/\{y_j\}} [\Phi_{y_i}(x_i;\bar{\zeta})-\Phi_{c}(x_i;\bar{\zeta})].
\end{equation}
Therefore, the value of $\alpha$ depends on the entire training dataset, which is not accessible. To avoid the accumulation of errors in estimating $q_{min}$ using generated samples during training, we directly optimize $\alpha$ as a trainable parameter.

\textbf{An extension to multiple classifiers. }
We have previously discussed a method using a single classifier to train a generator. Here, we present an extension of our approach to employ multiple classifiers for training a single generator. Given $T$ classifiers $\{\Phi^{(t)}\}_{t\in[T]}$, we encode the training data information from distinct classifiers into one generator by incorporating the classifier index $t$ as an input of the generator. By providing random noise $\epsilon$, label $y$, and classifier index $t$ as inputs, $\bm{x}$ is generated as $\bm{x}=g(\epsilon, y, t; \theta)$. During the training process, we compute the loss for each classifier, $L^{(t)}(\theta, \eta, \alpha^{(t)})$, and aggregate them as the final optimization objective. For each classifier, we optimized a distinct $\alpha$ and use $\alpha^{(t)}$ to denote the $\alpha$ for the $t$-th classifier. We utilize a unified network $h: \mathcal{R}^d \times \mathcal{Y} \times [T] \rightarrow \mathcal{R}^{|\mathcal{Y}|}$ to compute the KKT multipliers, which also includes the classifier index as an input. The optimization objective for the training process can be formulated as:
\begin{equation}
    \min_{\theta, \eta, \{\alpha^{(t)}\}_{t\in[T]}} \sum_{t\in[T]}L^{(t)}(\theta, \eta, \alpha^{(t)}).
\end{equation}
Upon completion of training, generating samples requires an additional step to sample a suitable classifier index. Let $\mathcal{T}_{y}$ denote the set of classifier indices, where the indices in the set correspond to classifiers whose training set contains samples with label $y$. To generate a sample $\bm{x}$ with label $y$, we first sample a classifier index $t$ from $\mathcal{T}_{y}$, and then utilize $t$, along with $y$ and random noise $\epsilon$, as inputs to the generator to produce the sample.
\section{Experiments}
\subsection{An Example of 2D Synthetic Data}
\begin{figure}
  \centering
  \begin{subfigure}{0.31\linewidth}
    \includegraphics[width=\linewidth]{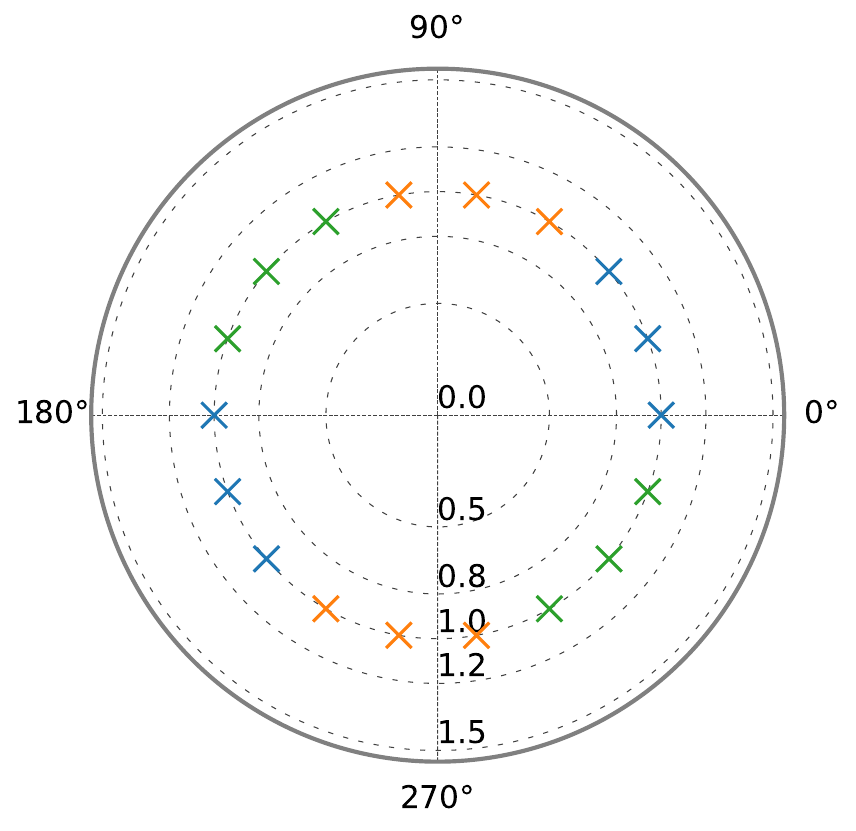}
    \caption{Training Data}
    \label{fig:toy:one:a}
  \end{subfigure}
  \begin{subfigure}{0.31\linewidth}
    \includegraphics[width=\linewidth]{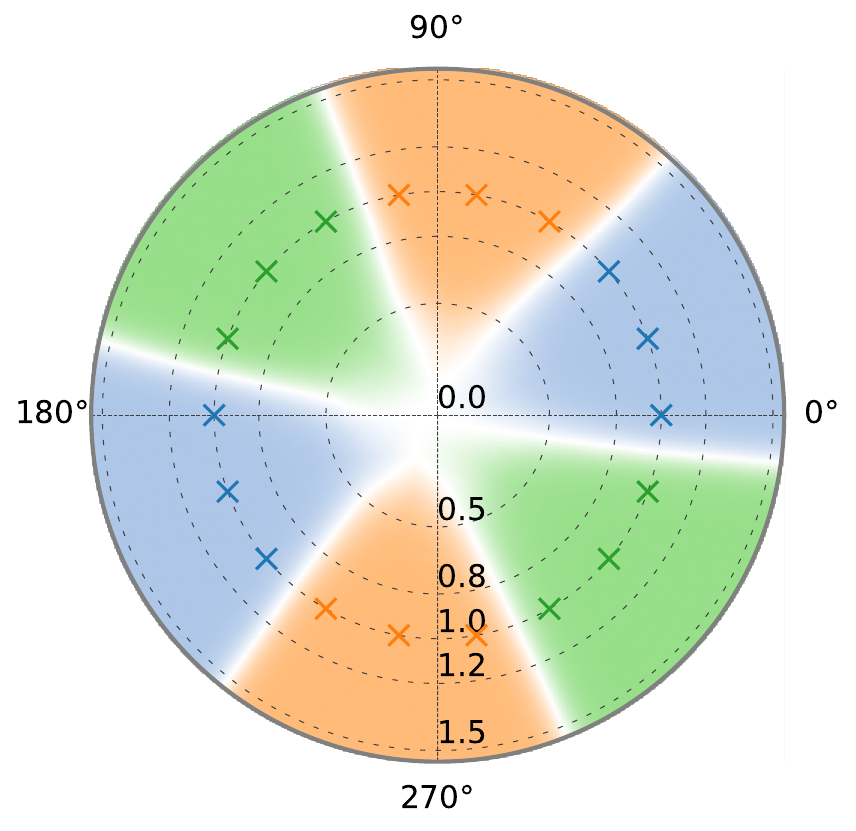}
    \caption{Prediction Landscape}
    \label{fig:toy:one:b}
  \end{subfigure}
  \begin{subfigure}{0.31\linewidth}
    \includegraphics[width=\linewidth]{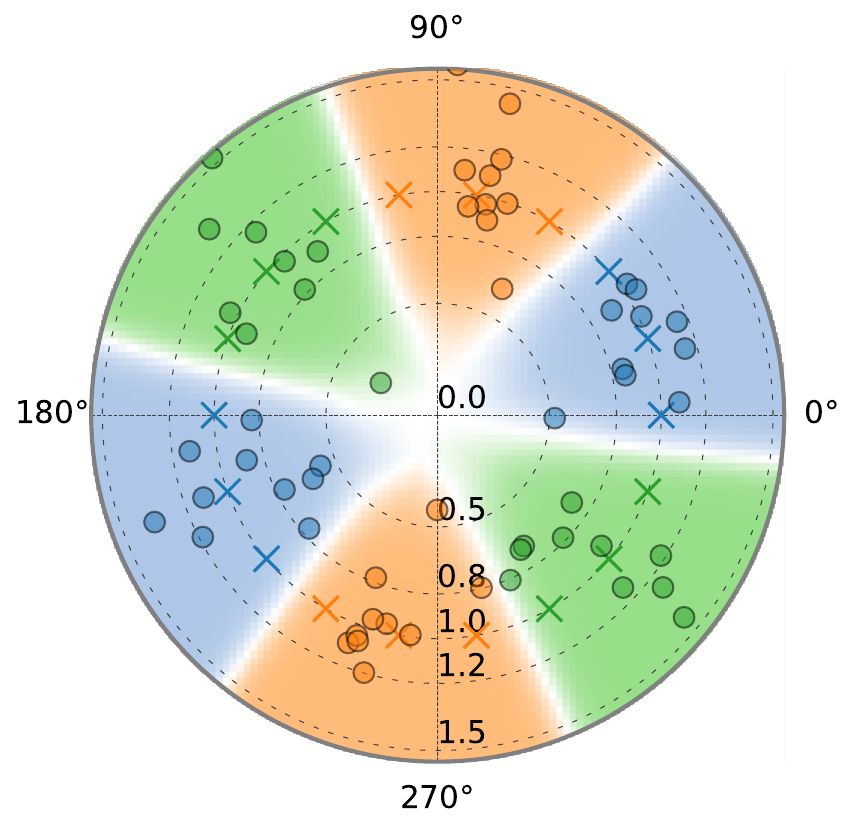}
    \caption{Generated Data}
    \label{fig:toy:one:c}
  \end{subfigure}
  \caption{An illustrative 2D example showcases our proposed method. \cref{fig:toy:one:a,fig:toy:one:b,fig:toy:one:c} are the training data employed for the classifier, the classifier's learned prediction landscape, and the generated samples using the learned generator, respectively.}
  \label{fig:toy:one}\vspace{-1em}
\end{figure}
\begin{figure}
\centering
\tabskip=0pt
\arrayrulecolor{gray!50}
\begin{tabular}{|c@{}c@{}c|c@{}c|}
\hline
    \hspace{-0.5em}
  \begin{subfigure}{.18\textwidth}
  \centering
  \includegraphics[width=\textwidth]{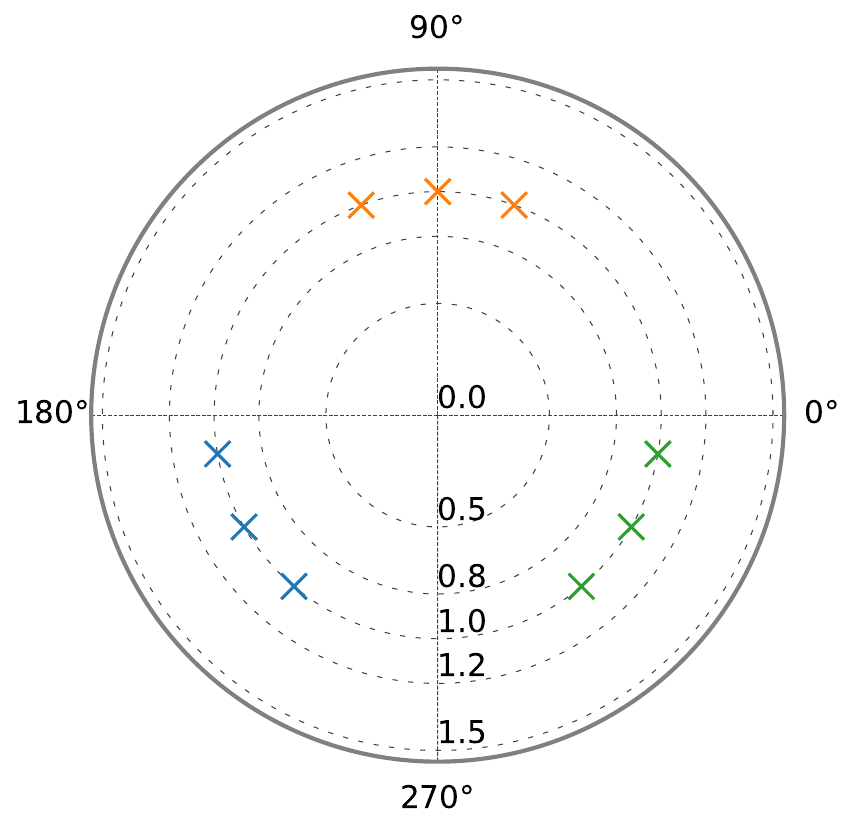}
  \caption{}\label{fig:toy:two:a}
  \end{subfigure} &
  \begin{subfigure}{.18\textwidth}
  \centering
  \includegraphics[width=\textwidth]{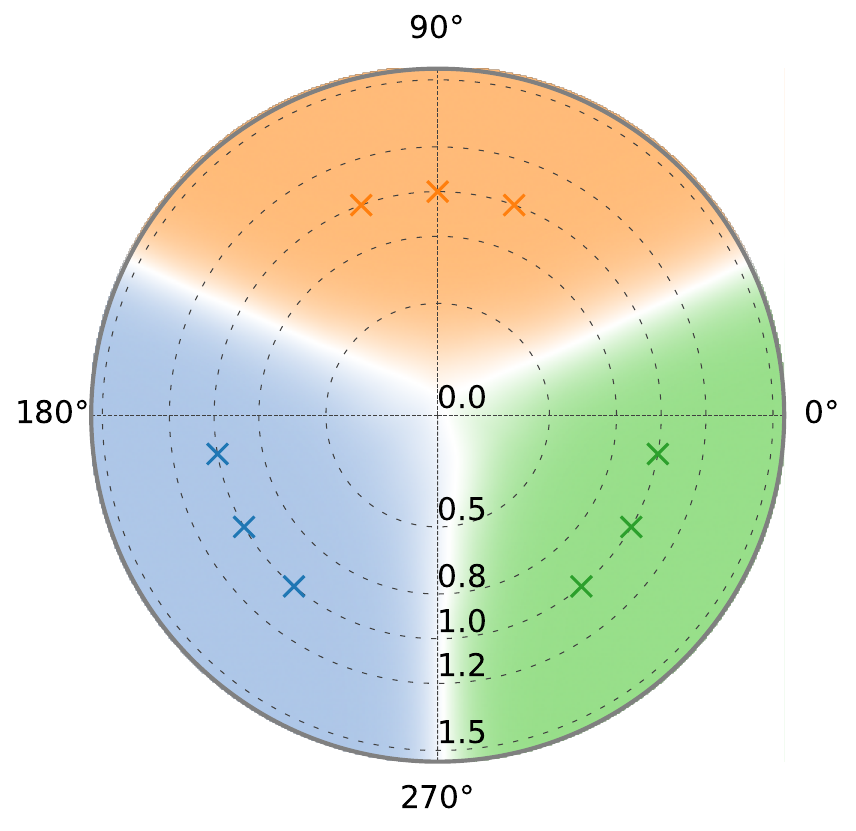}
  \caption{}\label{fig:toy:two:b}
  \end{subfigure} &
  \begin{subfigure}{.18\textwidth}
  \centering
  \includegraphics[width=\textwidth]{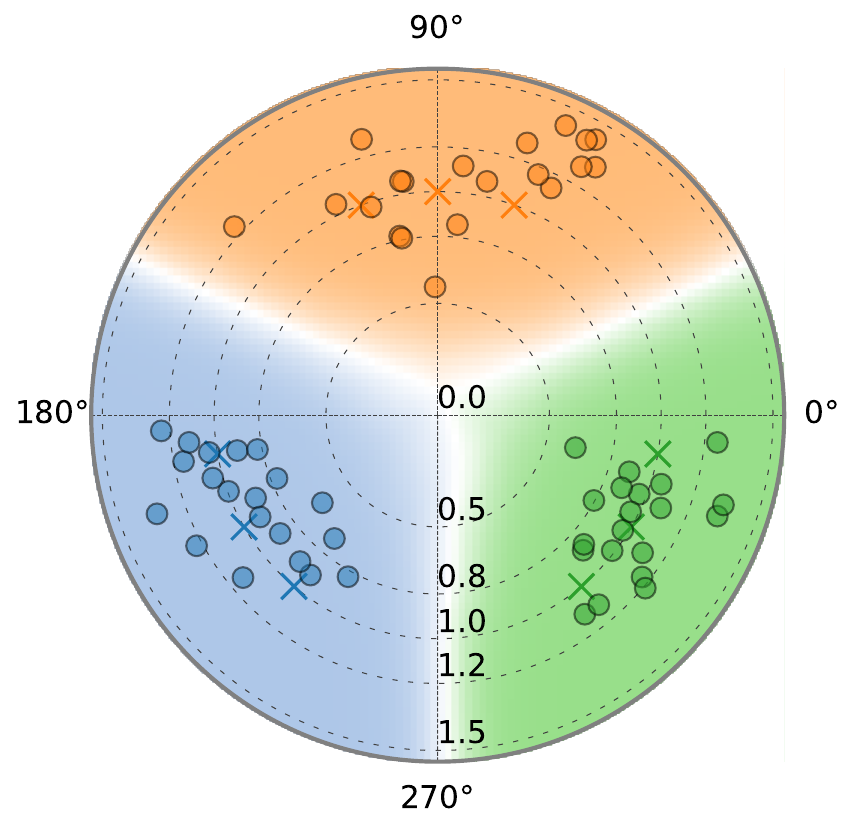}
  \caption{}\label{fig:toy:two:c}
  \end{subfigure} &
  \multirow{2}{*}{
    \hspace{-1em}
    \begin{subfigure}{.25\textwidth}
    \centering
    \vspace{-3.7em}
    \includegraphics[width=\textwidth]{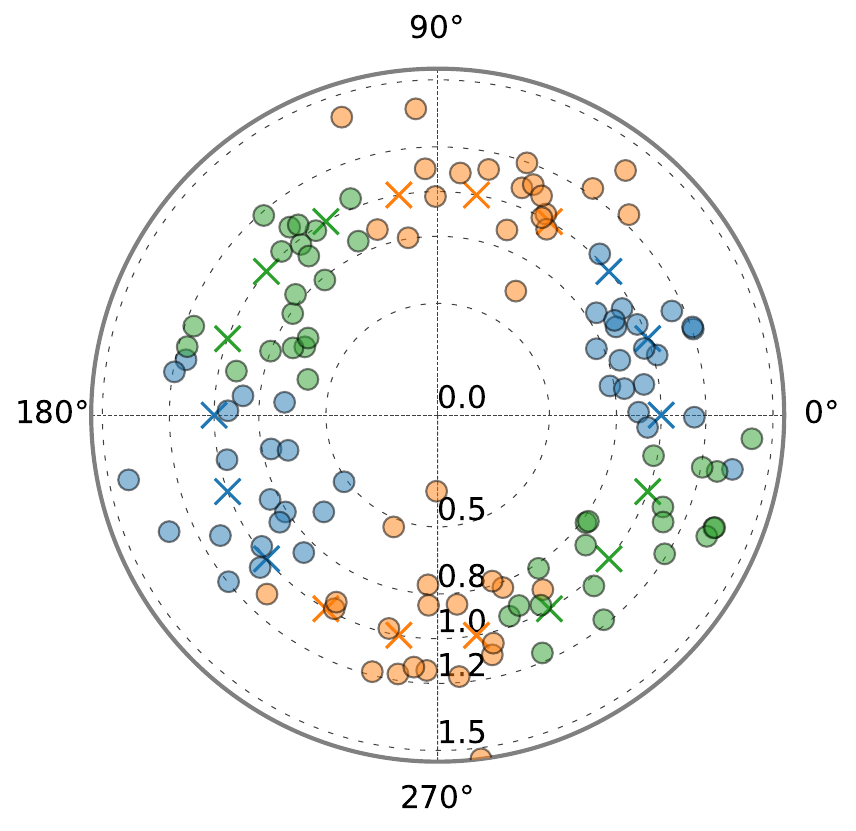}
    \caption{}\label{fig:toy:two:d}
    \end{subfigure}
    \hspace{-1em}
  } &
  \begin{subfigure}{.18\textwidth}
  \centering
  \includegraphics[width=\textwidth]{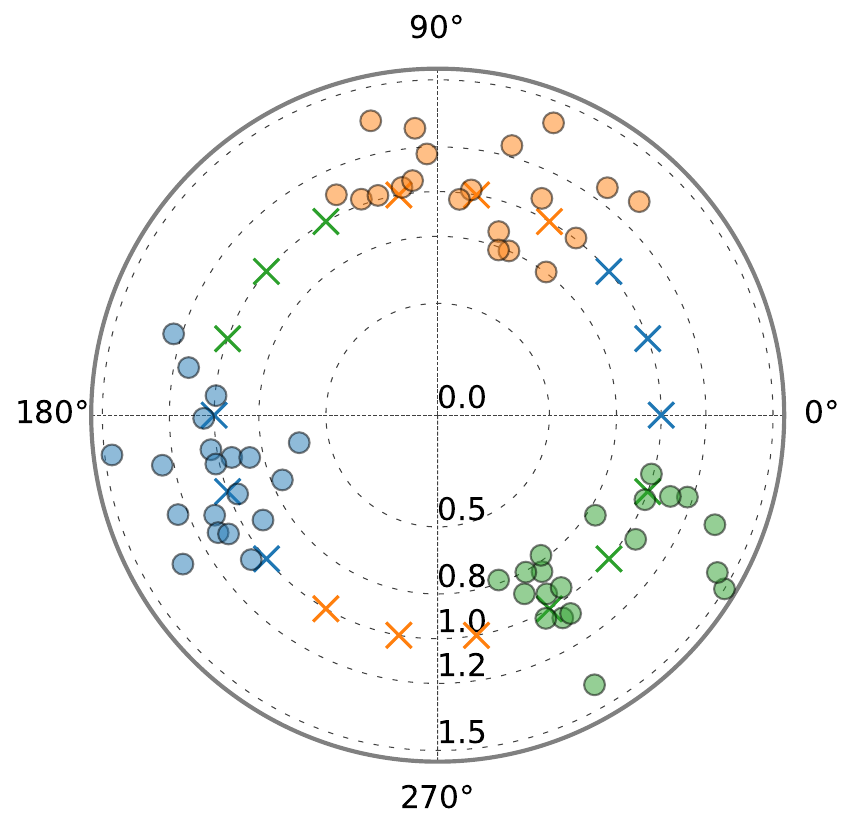}
  \caption{}\label{fig:toy:two:e}
  \end{subfigure}\hspace{-0.5em}\\\cline{1-3}
  \hspace{-0.5em}
\begin{subfigure}{.18\textwidth}
  \centering
  \vspace{0.1em}
  \includegraphics[width=\textwidth]{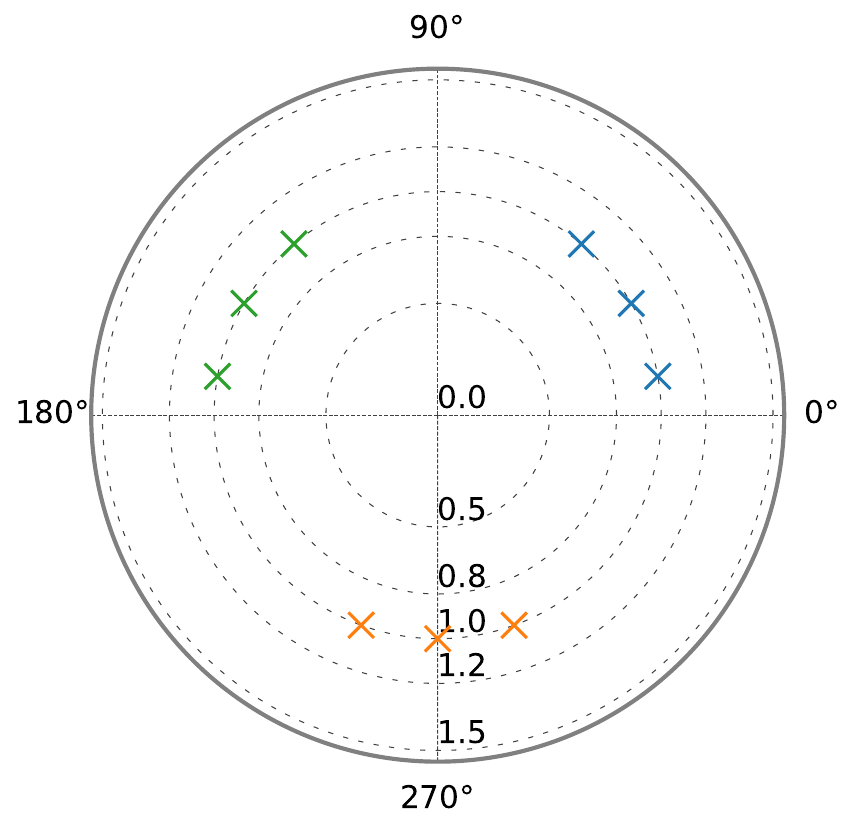}
  \caption{}\label{fig:toy:two:f}
  \end{subfigure} &
  \begin{subfigure}{.18\textwidth}
  \centering
  \vspace{0.1em}
  \includegraphics[width=\textwidth]{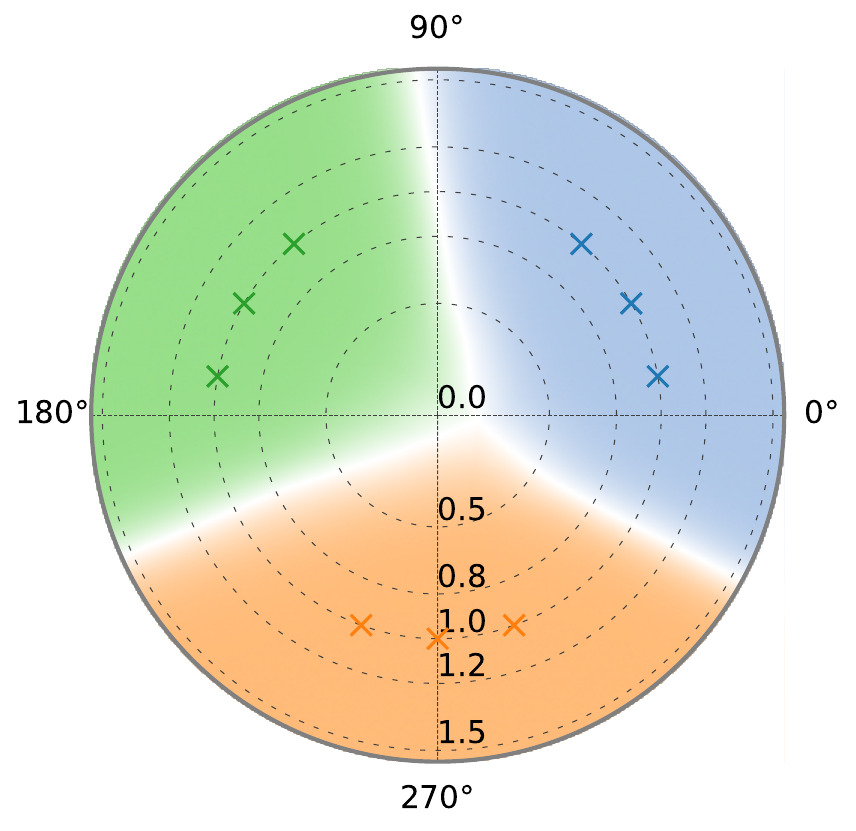}
  \caption{}\label{fig:toy:two:g}
  \end{subfigure} &
  \begin{subfigure}{.18\textwidth}
  \centering
  \vspace{0.1em}
  \includegraphics[width=\textwidth]{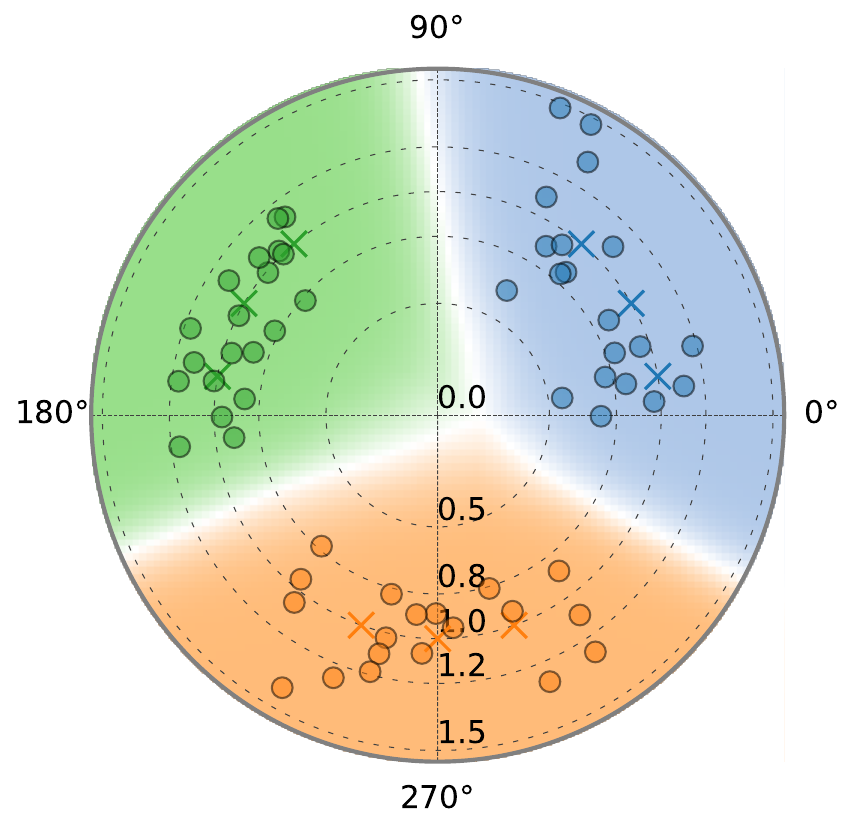}
  \caption{}\label{fig:toy:two:h}
  \end{subfigure} &
  &
  \begin{subfigure}{.18\textwidth}
  \centering
  \vspace{0.1em}
  \includegraphics[width=\textwidth]{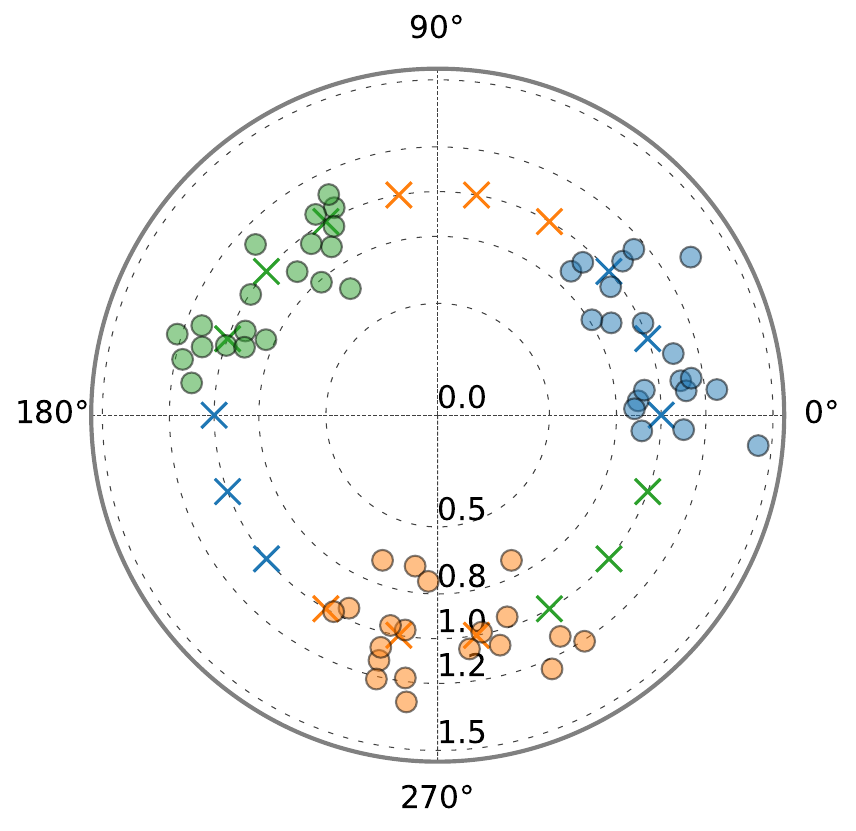}
  \caption{}\label{fig:toy:two:i}
  \end{subfigure}\hspace{-0.5em}\\
  \hline
\end{tabular}

\caption{An illustrative 2D example showcases the process of using two pre-trained classifiers to train a generator. \cref{fig:toy:two:a,fig:toy:two:b,fig:toy:two:c} and \cref{fig:toy:two:f,fig:toy:two:g,fig:toy:two:h} are two groups of training data, classifier's learned prediction landscape, and generator's generated samples. \cref{fig:toy:two:d} shows the generated samples of the generator trained using two classifiers. \cref{fig:toy:two:e,fig:toy:two:i} are the generated samples of the generator in \cref{fig:toy:two:d} with fixed classifier index.}
    \label{fig:toy:two}
\end{figure}
In this subsection, we employ a two-dimensional example to analyze the proposed approach. We evenly generate 18 data points on a unit circle, and categorize them into three classes to create our training dataset $D$. The distribution of $D$ is plotted in \cref{fig:toy:one:a}, with distinct colors representing different categories. Utilizing this data, we train a three-layer fully connected network $\Phi$ as the classifier. The prediction landscape of $\Phi$ is displayed in \cref{fig:toy:one:b}. The color of the region represents the category predicted by the classifier for samples within that region. The white areas represent the classifier's decision boundary. Given $\Phi$, we train a generator $g$ using the proposed method. The samples generated by $g$ are plotted in \cref{fig:toy:one:c}. Despite the presence of certain noise, the distribution of the generated data aligns consistently with that of the classifier's training data.

Leveraging this two-dimensional example, we further analyze our proposed method of training a generator using multiple classifiers. We evenly split the previous training dataset into $D_1$ and $D_2$, plotted in \cref{fig:toy:two:a,fig:toy:two:f}, respectively. Using these two subsets of data, we train two classifiers, $\Phi_1$ and $\Phi_2$, and subsequently train two generators, $g_1$ and $g_2$. \cref{fig:toy:two:c,fig:toy:two:h} display the data generated by $g_1$ and $g_2$, respectively. Here, during the training of $g_1$, we solely use $\Phi_1$, and during the training of $g_2$, we solely use $\Phi_2$. Obviously in \cref{fig:toy:one:b,fig:toy:two:g}, owing to the generalization ability of neural networks, $\Phi_1$ and $\Phi_2$ learn additional categorization capabilities beyond the training data, with prediction areas with the same color surpassing the region covered by the training data. Contrarily, the samples generated by training $g_1$ and $g_2$ congregate around the actual data points, effectively recovering the original distribution of the training data. Subsequently, by employing the extension of the proposed method, we train a generator $g_{1+2}$ using $\Phi_1$ and $\Phi_2$ together. As shown in \cref{fig:toy:two:d}, $g_{1+2}$ possesses the generative capabilities of both $g_1$ and $g_2$, capable of generating samples on the entire training data $D$. We further fix the classifier index of $g_{1+2}$ to either $1$ or $2$, randomly sample noise and category labels, and observe the data generated by $g_{1+2}$. As depicted in \cref{fig:toy:two:e,fig:toy:two:h}, $g_{1+2}$  is also capable of independently generating data belonging to either $D_1$ or $D_2$.

\subsection{Image Generation}
\begin{figure}
  \centering
    \includegraphics[width=\linewidth]{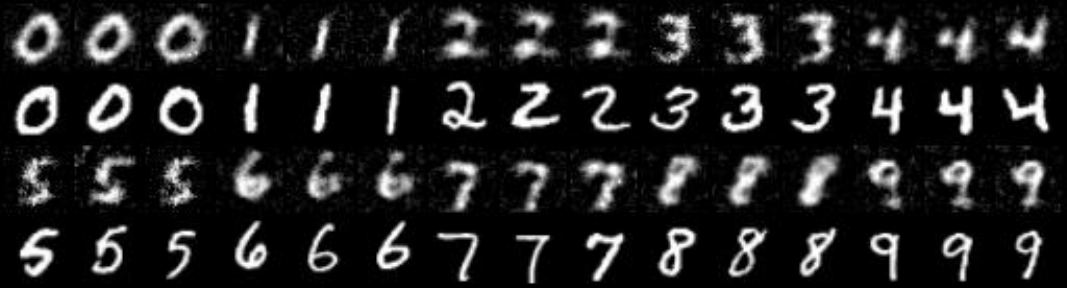}
    \caption{Generator-produced samples. The generator is trained using a single classifier trained on the MNIST dataset.}
  \label{fig:mnist}
\end{figure}
In this subsection, we showcase the experimental results on the MNIST~\citep{mnist} and CelebA~\citep{celeba} datasets. More results and implementation details are left in the appendix. For the MNIST dataset, we set up a classification task corresponding to the digits 0-9 with 500 training data (50 images per class) randomly sampled from the original training set. For the CelebA dataset, we utilized various binary attributes to construct binary classification tasks on facial images, for example, distinguishing between males and females. For each task, we randomly sampled 100 images (50 images per class) from the original training dataset and resize them to 32x32 to be our training dataset.

We employed a three-layer fully-connected network with a ReLU activation function and batch normalization as the classifier for the aforementioned classification tasks. 
The networks were trained until the classification loss converges using full batch gradient descent, which ensures the parameters  are close to the convergence point required in the theory of Maximum-Margin Bias.

We use generators composed of three fully-connected layers followed by three transposed convolution layers, with ReLU activation function and batch normalization. Network parameters were initialized using Kaiming initialization~\citep{heinit} and trained for $50,000$ epochs. The batch size and learning rate were set as hyperparameters and optimized via random search. In order to control the noise in generated images, we also utilized total variation loss in pixel space as a regularization term.

\begin{figure}
  \centering
  \begin{subfigure}{0.32\linewidth}
    \includegraphics[width=\linewidth]{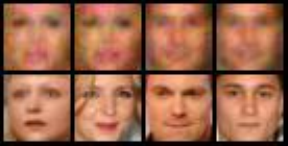}
    \caption{Gender}
    \label{fig:celeba:a}
  \end{subfigure}
  \begin{subfigure}{0.32\linewidth}
    \includegraphics[width=\linewidth]{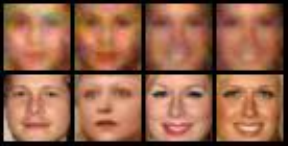}
    \caption{Mouth Opened}
    \label{fig:celeba:b}
  \end{subfigure}
  \begin{subfigure}{0.32\linewidth}
    \includegraphics[width=\linewidth]{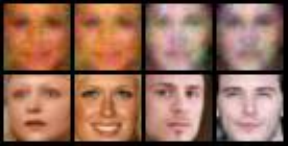}
    \caption{Pale Skin}
    \label{fig:celeba:c}
  \end{subfigure}
  \caption{Generator-produced samples. The generator is trained using a single classifier trained on the CelebA dataset. The captions of the subfigures indicate the attributes used as the label.}
  \label{fig:celeba}\vspace{-0.5em}
\end{figure}

The generated results on MNIST and CelebA are shown in \cref{fig:mnist,fig:celeba}. Odd rows present the generated images, and even rows present the images from the training dataset that are closest to the generated images. The distance between images is measured by the SSIM metric. As shown by the results, the generator trained using our method is capable of generating digits and facial images, even though it has never been exposed to images of digits or faces.

\begin{figure}
  \centering
    \includegraphics[width=\linewidth]{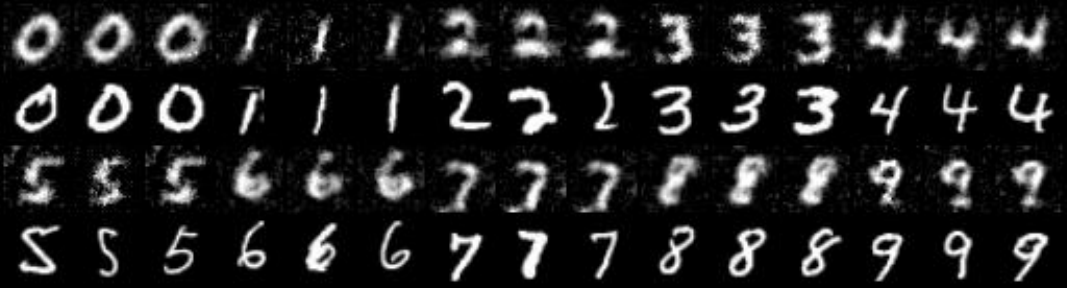}
    \caption{Generator-produced samples. The generator is trained using two classifiers trained on the MNIST dataset.}
  \label{fig:mnist_en}\vspace{-1em}
\end{figure}

To validate the extended method we proposed for multiple classifiers, we partitioned the aforementioned digit classification dataset into two subsets including digits $0-4$ and digits $5-9$, respectively, and trained two classifiers separately. We then employed our method to train a single generator using both classifiers. \cref{fig:mnist_en} showcases the final images generated. The trained generator successfully integrates information from both classifiers, being capable of generating all digits from 0-9.

\section{Conclusion}
In this research, we investigate a pioneering task: training a generator directly utilizing a pre-trained classifier, devoid of training data. Based on the maximum margin bias theory, we present the relationship between pre-trained neural network parameters and the training data distribution. Consequently, we devise an innovative loss function to enable the generator's training. Essentially, our loss function requires the generator to guarantee the optimality of the parameters of the pre-trained classifier under its generated data distribution. 
From a broader perspective, the reuse and revision of pre-trained neural networks have been a widely studied direction.~\citep{lab1,lab2,lab3,lab4,nreuse2,nreuse1} Our method offers a novel direction for leveraging pre-trained models.

\section*{Acknowledgment}
This project is supported by the National Research Foundation, 
Singapore under its AI Singapore Programme (AISG Award No: AISG2-RP-2021-023),
and the Singapore Ministry of Education Academic Research Fund Tier 1 (WBS: A0009440-01-00).

\clearpage
{\small
\bibliographystyle{plainnat}
\bibliography{main}

\begin{thebibliography}{64}
\providecommand{\natexlab}[1]{#1}
\providecommand{\url}[1]{\texttt{#1}}
\expandafter\ifx\csname urlstyle\endcsname\relax
  \providecommand{\doi}[1]{doi: #1}\else
  \providecommand{\doi}{doi: \begingroup \urlstyle{rm}\Url}\fi

\bibitem[Arjovsky et~al.(2017)Arjovsky, Chintala, and Bottou]{wgan}
Mart{\'{\i}}n Arjovsky, Soumith Chintala, and L{\'{e}}on Bottou.
\newblock Wasserstein generative adversarial networks.
\newblock In \emph{International Conference on Machine Learning (ICML)}, 2017.

\bibitem[Arpit et~al.(2017)Arpit, Jastrzebski, Ballas, Krueger, Bengio, Kanwal, Maharaj, Fischer, Courville, Bengio, and Lacoste{-}Julien]{memorization_3}
Devansh Arpit, Stanislaw Jastrzebski, Nicolas Ballas, David Krueger, Emmanuel Bengio, Maxinder~S. Kanwal, Tegan Maharaj, Asja Fischer, Aaron~C. Courville, Yoshua Bengio, and Simon Lacoste{-}Julien.
\newblock A closer look at memorization in deep networks.
\newblock In \emph{International Conference on Machine Learning (ICML)}, 2017.

\bibitem[Brock et~al.(2019)Brock, Donahue, and Simonyan]{biggan}
Andrew Brock, Jeff Donahue, and Karen Simonyan.
\newblock Large scale {GAN} training for high fidelity natural image synthesis.
\newblock In \emph{International Conference on Learning Representations (ICLR)}, 2019.

\bibitem[Chen et~al.(2016)Chen, Duan, Houthooft, Schulman, Sutskever, and Abbeel]{infogan}
Xi~Chen, Yan Duan, Rein Houthooft, John Schulman, Ilya Sutskever, and Pieter Abbeel.
\newblock Infogan: Interpretable representation learning by information maximizing generative adversarial nets.
\newblock In \emph{Conference on Neural Information Processing Systems (NeurlPS)}, 2016.

\bibitem[Daniely(2020)]{memorization_2}
Amit Daniely.
\newblock Neural networks learning and memorization with (almost) no over-parameterization.
\newblock In \emph{Conference on Neural Information Processing Systems (NeurlPS)}, 2020.

\bibitem[Davis et~al.(2020)Davis, Drusvyatskiy, Kakade, and Lee]{DavisDKL20}
Damek Davis, Dmitriy Drusvyatskiy, Sham~M. Kakade, and Jason~D. Lee.
\newblock Stochastic subgradient method converges on tame functions.
\newblock \emph{Found. Comput. Math.}, 20\penalty0 (1):\penalty0 119--154, 2020.

\bibitem[Denton et~al.(2015)Denton, Chintala, Szlam, and Fergus]{lagan}
Emily~L. Denton, Soumith Chintala, Arthur Szlam, and Rob Fergus.
\newblock Deep generative image models using a laplacian pyramid of adversarial networks.
\newblock In \emph{Conference on Neural Information Processing Systems (NeurlPS)}, 2015.

\bibitem[Donahue et~al.(2017)Donahue, Kr{\"{a}}henb{\"{u}}hl, and Darrell]{bigan}
Jeff Donahue, Philipp Kr{\"{a}}henb{\"{u}}hl, and Trevor Darrell.
\newblock Adversarial feature learning.
\newblock In \emph{International Conference on Learning Representations (ICLR)}, 2017.

\bibitem[Dosovitskiy et~al.(2021)Dosovitskiy, Beyer, Kolesnikov, Weissenborn, Zhai, Unterthiner, Dehghani, Minderer, Heigold, Gelly, Uszkoreit, and Houlsby]{vit}
Alexey Dosovitskiy, Lucas Beyer, Alexander Kolesnikov, Dirk Weissenborn, Xiaohua Zhai, Thomas Unterthiner, Mostafa Dehghani, Matthias Minderer, Georg Heigold, Sylvain Gelly, Jakob Uszkoreit, and Neil Houlsby.
\newblock An image is worth 16x16 words: Transformers for image recognition at scale.
\newblock In \emph{International Conference on Learning Representations (ICLR)}, 2021.

\bibitem[Engstrom et~al.(2019)Engstrom, Ilyas, Santurkar, Tsipras, Tran, and Madry]{fv_1}
Logan Engstrom, Andrew Ilyas, Shibani Santurkar, Dimitris Tsipras, Brandon Tran, and Aleksander Madry.
\newblock Adversarial robustness as a prior for learned representations, 2019.

\bibitem[Fang et~al.(2023{\natexlab{a}})Fang, Ma, Song, Mi, and Wang]{lab2}
Gongfan Fang, Xinyin Ma, Mingli Song, Michael~Bi Mi, and Xinchao Wang.
\newblock Depgraph: Towards any structural pruning.
\newblock In \emph{IEEE / CVF Computer Vision and Pattern Recognition Conference (CVPR)}, 2023{\natexlab{a}}.

\bibitem[Fang et~al.(2023{\natexlab{b}})Fang, Ma, and Wang]{lab3}
Gongfan Fang, Xinyin Ma, and Xinchao Wang.
\newblock Structural pruning for diffusion models.
\newblock In \emph{Conference on Neural Information Processing Systems (NeurlPS)}, 2023{\natexlab{b}}.

\bibitem[Feldman and Zhang(2020)]{memorization_1}
Vitaly Feldman and Chiyuan Zhang.
\newblock What neural networks memorize and why: Discovering the long tail via influence estimation.
\newblock In \emph{Conference on Neural Information Processing Systems (NeurlPS)}, 2020.

\bibitem[Fredrikson et~al.(2015)Fredrikson, Jha, and Ristenpart]{mi_attack_2}
Matt Fredrikson, Somesh Jha, and Thomas Ristenpart.
\newblock Model inversion attacks that exploit confidence information and basic countermeasures.
\newblock In \emph{Conference on Computer and Communications Security}, 2015.

\bibitem[Goodfellow et~al.(2014)Goodfellow, Pouget{-}Abadie, Mirza, Xu, Warde{-}Farley, Ozair, Courville, and Bengio]{gan}
Ian~J. Goodfellow, Jean Pouget{-}Abadie, Mehdi Mirza, Bing Xu, David Warde{-}Farley, Sherjil Ozair, Aaron~C. Courville, and Yoshua Bengio.
\newblock Generative adversarial nets.
\newblock In \emph{Conference on Neural Information Processing Systems (NeurlPS)}, 2014.

\bibitem[Grathwohl et~al.(2020)Grathwohl, Wang, Jacobsen, Duvenaud, Norouzi, and Swersky]{generative_classifier_joint_train_1}
Will Grathwohl, Kuan{-}Chieh Wang, J{\"{o}}rn{-}Henrik Jacobsen, David Duvenaud, Mohammad Norouzi, and Kevin Swersky.
\newblock Your classifier is secretly an energy based model and you should treat it like one.
\newblock In \emph{International Conference on Learning Representations (ICLR)}, 2020.

\bibitem[Guo et~al.(2023)Guo, Ma, Jiang, Yuan, Yu, and Luo]{generative_classifier_joint_train_2}
Qiushan Guo, Chuofan Ma, Yi~Jiang, Zehuan Yuan, Yizhou Yu, and Ping Luo.
\newblock Egc: Image generation and classification via a single energy-based model.
\newblock \emph{arXiv}, 2023.

\bibitem[Haim et~al.(2022)Haim, Vardi, Yehudai, Shamir, and Irani]{DataRec}
Niv Haim, Gal Vardi, Gilad Yehudai, Ohad Shamir, and Michal Irani.
\newblock Reconstructing training data from trained neural networks.
\newblock In \emph{Conference on Neural Information Processing Systems (NeurlPS)}, 2022.

\bibitem[He et~al.(2015)He, Zhang, Ren, and Sun]{heinit}
Kaiming He, Xiangyu Zhang, Shaoqing Ren, and Jian Sun.
\newblock Delving deep into rectifiers: Surpassing human-level performance on imagenet classification.
\newblock 2015.

\bibitem[He et~al.(2019)He, Zhang, and Lee]{mi_attack}
Zecheng He, Tianwei Zhang, and Ruby~B. Lee.
\newblock Model inversion attacks against collaborative inference.
\newblock In David Balenson, editor, \emph{Proceedings of the 35th Annual Computer Security Applications Conference, {ACSAC}}, 2019.

\bibitem[Isola et~al.(2017)Isola, Zhu, Zhou, and Efros]{pix2pix}
Phillip Isola, Jun{-}Yan Zhu, Tinghui Zhou, and Alexei~A. Efros.
\newblock Image-to-image translation with conditional adversarial networks.
\newblock In \emph{IEEE / CVF Computer Vision and Pattern Recognition Conference (CVPR)}, 2017.

\bibitem[Jeon et~al.(2021)Jeon, Kim, Lee, Oh, and Ok]{mi_gan_1}
Jinwoo Jeon, Jaechang Kim, Kangwook Lee, Sewoong Oh, and Jungseul Ok.
\newblock Gradient inversion with generative image prior.
\newblock In \emph{Conference on Neural Information Processing Systems (NeurlPS)}, 2021.

\bibitem[Ji and Telgarsky(2020)]{mm_homo_5}
Ziwei Ji and Matus Telgarsky.
\newblock Directional convergence and alignment in deep learning.
\newblock In \emph{Conference on Neural Information Processing Systems (NeurlPS)}, 2020.

\bibitem[Jing et~al.(2023)Jing, Yuan, Ju, Yang, Wang, and Tao]{jing2023deep}
Yongcheng Jing, Chongbin Yuan, Li~Ju, Yiding Yang, Xinchao Wang, and Dacheng Tao.
\newblock Deep graph reprogramming.
\newblock In \emph{CVPR}, 2023.

\bibitem[Karras et~al.(2018)Karras, Aila, Laine, and Lehtinen]{progan}
Tero Karras, Timo Aila, Samuli Laine, and Jaakko Lehtinen.
\newblock Progressive growing of gans for improved quality, stability, and variation.
\newblock In \emph{International Conference on Learning Representations (ICLR)}, 2018.

\bibitem[Kunin et~al.(2023)Kunin, Yamamura, Ma, and Ganguli]{mmb}
Daniel Kunin, Atsushi Yamamura, Chao Ma, and Surya Ganguli.
\newblock The asymmetric maximum margin bias of quasi-homogeneous neural networks.
\newblock \emph{International Conference on Learning Representations (ICLR)}, 2023.

\bibitem[Le and Jegelka(2022)]{mm_homo_6}
Thien Le and Stefanie Jegelka.
\newblock Training invariances and the low-rank phenomenon: beyond linear networks.
\newblock In \emph{International Conference on Learning Representations (ICLR)}, 2022.

\bibitem[Lecun et~al.(1998)Lecun, Bottou, Bengio, and Haffner]{mnist}
Y.~Lecun, L.~Bottou, Y.~Bengio, and P.~Haffner.
\newblock Gradient-based learning applied to document recognition.
\newblock \emph{Proceedings of the IEEE}, 1998.

\bibitem[LeCun et~al.(2006)LeCun, Chopra, Hadsell, Ranzato, and Huang]{classifier_energy_based_model}
Yann LeCun, Sumit Chopra, Raia Hadsell, M~Ranzato, and Fujie Huang.
\newblock A tutorial on energy-based learning.
\newblock \emph{Predicting structured data}, 1\penalty0 (0), 2006.

\bibitem[Ledig et~al.(2017)Ledig, Theis, Huszar, Caballero, Cunningham, Acosta, Aitken, Tejani, Totz, Wang, and Shi]{srgan}
Christian Ledig, Lucas Theis, Ferenc Huszar, Jose Caballero, Andrew Cunningham, Alejandro Acosta, Andrew~P. Aitken, Alykhan Tejani, Johannes Totz, Zehan Wang, and Wenzhe Shi.
\newblock Photo-realistic single image super-resolution using a generative adversarial network.
\newblock In \emph{IEEE / CVF Computer Vision and Pattern Recognition Conference (CVPR)}, 2017.

\bibitem[Linh et~al.(2020)Linh, Nguyen, and Arai]{gan_denoising-1}
Tran~Duy Linh, Son~Minh Nguyen, and Masayuki Arai.
\newblock Gan-based noise model for denoising real images.
\newblock In \emph{Computer Vision - {ACCV} 2020 - 15th Asian Conference on Computer Vision, Kyoto, Japan, November 30 - December 4, 2020, Revised Selected Papers, Part {IV}}, 2020.

\bibitem[Liu et~al.(2015)Liu, Luo, Wang, and Tang]{celeba}
Ziwei Liu, Ping Luo, Xiaogang Wang, and Xiaoou Tang.
\newblock Deep learning face attributes in the wild.
\newblock In \emph{International Conference on Computer Vision (ICCV)}, December 2015.

\bibitem[Lyu and Li(2020{\natexlab{a}})]{LyuL20}
Kaifeng Lyu and Jian Li.
\newblock Gradient descent maximizes the margin of homogeneous neural networks.
\newblock In \emph{International Conference on Learning Representations (ICLR)}, 2020{\natexlab{a}}.

\bibitem[Lyu and Li(2020{\natexlab{b}})]{mm_homo_4}
Kaifeng Lyu and Jian Li.
\newblock Gradient descent maximizes the margin of homogeneous neural networks.
\newblock In \emph{International Conference on Learning Representations (ICLR)}, 2020{\natexlab{b}}.

\bibitem[Ma et~al.(2023)Ma, Fang, and Wang]{lab1}
Xinyin Ma, Gongfan Fang, and Xinchao Wang.
\newblock {LLM-Pruner: On the Structural Pruning of Large Language Models}.
\newblock In \emph{Conference on Neural Information Processing Systems (NeurlPS)}, 2023.

\bibitem[Mirza and Osindero(2014)]{cgan}
Mehdi Mirza and Simon Osindero.
\newblock Conditional generative adversarial nets.
\newblock \emph{CoRR}, 2014.

\bibitem[Nacson et~al.(2019)Nacson, Gunasekar, Lee, Srebro, and Soudry]{mm_homo_3}
Mor~Shpigel Nacson, Suriya Gunasekar, Jason~D. Lee, Nathan Srebro, and Daniel Soudry.
\newblock Lexicographic and depth-sensitive margins in homogeneous and non-homogeneous deep models.
\newblock In \emph{International Conference on Machine Learning (ICML)}, 2019.

\bibitem[Nguyen et~al.(2017)Nguyen, Clune, Bengio, Dosovitskiy, and Yosinski]{fv_gan_1}
Anh Nguyen, Jeff Clune, Yoshua Bengio, Alexey Dosovitskiy, and Jason Yosinski.
\newblock Plug {\&} play generative networks: Conditional iterative generation of images in latent space.
\newblock In \emph{IEEE / CVF Computer Vision and Pattern Recognition Conference (CVPR)}, 2017.

\bibitem[Nguyen et~al.(2016)Nguyen, Yosinski, and Clune]{fv_3}
Anh~Mai Nguyen, Jason Yosinski, and Jeff Clune.
\newblock Multifaceted feature visualization: Uncovering the different types of features learned by each neuron in deep neural networks.
\newblock \emph{CoRR}, 2016.

\bibitem[Nowozin et~al.(2016)Nowozin, Cseke, and Tomioka]{fgan}
Sebastian Nowozin, Botond Cseke, and Ryota Tomioka.
\newblock f-gan: Training generative neural samplers using variational divergence minimization.
\newblock In \emph{Conference on Neural Information Processing Systems (NeurlPS)}, 2016.

\bibitem[Odena et~al.(2017)Odena, Olah, and Shlens]{acgan}
Augustus Odena, Christopher Olah, and Jonathon Shlens.
\newblock Conditional image synthesis with auxiliary classifier gans.
\newblock In \emph{International Conference on Machine Learning (ICML)}, 2017.

\bibitem[Olah et~al.(2017)Olah, Mordvintsev, and Schubert]{fv_2}
Chris Olah, Alexander Mordvintsev, and Ludwig Schubert.
\newblock Feature visualization.
\newblock \emph{Distill}, 2\penalty0 (11):\penalty0 e7, 2017.

\bibitem[Rosset et~al.(2003)Rosset, Zhu, and Hastie]{mm_logist_1}
Saharon Rosset, Ji~Zhu, and Trevor Hastie.
\newblock Margin maximizing loss functions.
\newblock In \emph{Conference on Neural Information Processing Systems (NeurlPS)}, 2003.

\bibitem[Soudry et~al.(2018)Soudry, Hoffer, Nacson, Gunasekar, and Srebro]{mm_logist_2}
Daniel Soudry, Elad Hoffer, Mor~Shpigel Nacson, Suriya Gunasekar, and Nathan Srebro.
\newblock The implicit bias of gradient descent on separable data.
\newblock \emph{J. Mach. Learn. Res.}, 2018.

\bibitem[Tulyakov et~al.(2018)Tulyakov, Liu, Yang, and Kautz]{gan_video_2}
Sergey Tulyakov, Ming{-}Yu Liu, Xiaodong Yang, and Jan Kautz.
\newblock Mocogan: Decomposing motion and content for video generation.
\newblock In \emph{IEEE / CVF Computer Vision and Pattern Recognition Conference (CVPR)}, 2018.

\bibitem[Vaswani et~al.(2023)Vaswani, Shazeer, Parmar, Uszkoreit, Jones, Gomez, Kaiser, and Polosukhin]{attention}
Ashish Vaswani, Noam Shazeer, Niki Parmar, Jakob Uszkoreit, Llion Jones, Aidan~N. Gomez, Lukasz Kaiser, and Illia Polosukhin.
\newblock Attention is all you need, 2023.

\bibitem[Vondrick et~al.(2016)Vondrick, Pirsiavash, and Torralba]{gan_video_1}
Carl Vondrick, Hamed Pirsiavash, and Antonio Torralba.
\newblock Generating videos with scene dynamics.
\newblock In \emph{Conference on Neural Information Processing Systems (NeurlPS)}, 2016.

\bibitem[Wei et~al.(2019)Wei, Lee, Liu, and Ma]{mm_homo_1}
Colin Wei, Jason~D. Lee, Qiang Liu, and Tengyu Ma.
\newblock Regularization matters: Generalization and optimization of neural nets v.s. their induced kernel.
\newblock In \emph{Conference on Neural Information Processing Systems (NeurlPS)}, 2019.

\bibitem[Wei et~al.(2018)Wei, Gong, Liu, Lu, and Wang]{wgangp}
Xiang Wei, Boqing Gong, Zixia Liu, Wei Lu, and Liqiang Wang.
\newblock Improving the improved training of wasserstein gans: {A} consistency term and its dual effect.
\newblock In \emph{International Conference on Learning Representations (ICLR)}, 2018.

\bibitem[Wu et~al.(2017)Wu, Zheng, Zhang, and Huang]{gan_blending_1}
Huikai Wu, Shuai Zheng, Junge Zhang, and Kaiqi Huang.
\newblock {GP-GAN:} towards realistic high-resolution image blending.
\newblock \emph{CoRR}, 2017.

\bibitem[Wu et~al.(2016)Wu, Zhang, Xue, Freeman, and Tenenbaum]{gan_3d_1}
Jiajun Wu, Chengkai Zhang, Tianfan Xue, Bill Freeman, and Josh Tenenbaum.
\newblock Learning a probabilistic latent space of object shapes via 3d generative-adversarial modeling.
\newblock In \emph{Conference on Neural Information Processing Systems (NeurlPS)}, 2016.

\bibitem[Xu et~al.(2018)Xu, Zhou, Ji, and Liang]{mm_homo_2}
Tengyu Xu, Yi~Zhou, Kaiyi Ji, and Yingbin Liang.
\newblock Convergence of {SGD} in learning relu models with separable data.
\newblock \emph{CoRR}, 2018.

\bibitem[Yang et~al.(2017{\natexlab{a}})Yang, Lu, Lin, Shechtman, Wang, and Li]{gan_inpainting_2}
Chao Yang, Xin Lu, Zhe Lin, Eli Shechtman, Oliver Wang, and Hao Li.
\newblock High-resolution image inpainting using multi-scale neural patch synthesis.
\newblock In \emph{IEEE / CVF Computer Vision and Pattern Recognition Conference (CVPR)}, 2017{\natexlab{a}}.

\bibitem[Yang et~al.(2017{\natexlab{b}})Yang, Kannan, Batra, and Parikh]{gan_imagegenerate_1}
Jianwei Yang, Anitha Kannan, Dhruv Batra, and Devi Parikh.
\newblock {LR-GAN:} layered recursive generative adversarial networks for image generation.
\newblock In \emph{International Conference on Learning Representations (ICLR)}, 2017{\natexlab{b}}.

\bibitem[Yang et~al.(2022{\natexlab{a}})Yang, Ye, and Wang]{nreuse2}
Xingyi Yang, Jingwen Ye, and Xinchao Wang.
\newblock Factorizing knowledge in neural networks.
\newblock In \emph{European Conference on Computer Vision (ECCV)}, 2022{\natexlab{a}}.

\bibitem[Yang et~al.(2022{\natexlab{b}})Yang, Zhou, Liu, Ye, and Wang]{nreuse1}
Xingyi Yang, Daquan Zhou, Songhua Liu, Jingwen Ye, and Xinchao Wang.
\newblock Deep model reassembly.
\newblock In \emph{Conference on Neural Information Processing Systems (NeurlPS)}, 2022{\natexlab{b}}.

\bibitem[Yang et~al.(2019)Yang, Zhang, Chang, and Liang]{mi_gan_2}
Ziqi Yang, Jiyi Zhang, Ee{-}Chien Chang, and Zhenkai Liang.
\newblock Neural network inversion in adversarial setting via background knowledge alignment.
\newblock In \emph{Conference on Computer and Communications Security, {CCS}}, 2019.

\bibitem[Yeh et~al.(2017)Yeh, Chen, Lim, Schwing, Hasegawa{-}Johnson, and Do]{gan_inpainting}
Raymond~A. Yeh, Chen Chen, Teck{-}Yian Lim, Alexander~G. Schwing, Mark Hasegawa{-}Johnson, and Minh~N. Do.
\newblock Semantic image inpainting with deep generative models.
\newblock In \emph{IEEE / CVF Computer Vision and Pattern Recognition Conference (CVPR)}, 2017.

\bibitem[Yin et~al.(2020)Yin, Molchanov, Alvarez, Li, Mallya, Hoiem, Jha, and Kautz]{mi_1}
Hongxu Yin, Pavlo Molchanov, Jose~M. Alvarez, Zhizhong Li, Arun Mallya, Derek Hoiem, Niraj~K. Jha, and Jan Kautz.
\newblock Dreaming to distill: Data-free knowledge transfer via deepinversion.
\newblock In \emph{IEEE / CVF Computer Vision and Pattern Recognition Conference (CVPR)}, 2020.

\bibitem[Yu et~al.(2018)Yu, Lin, Yang, Shen, Lu, and Huang]{gan_inpainting_3}
Jiahui Yu, Zhe Lin, Jimei Yang, Xiaohui Shen, Xin Lu, and Thomas~S. Huang.
\newblock Generative image inpainting with contextual attention.
\newblock In \emph{IEEE / CVF Computer Vision and Pattern Recognition Conference (CVPR)}, 2018.

\bibitem[Yu et~al.(2023)Yu, Liu, Yang, and Wang]{lab4}
Runpeng Yu, Songhua Liu, Xingyi Yang, and Xinchao Wang.
\newblock Distribution shift inversion for out-of-distribution prediction.
\newblock In \emph{IEEE / CVF Computer Vision and Pattern Recognition Conference (CVPR)}, 2023.

\bibitem[Zhang et~al.(2020)Zhang, Jia, Pei, Wang, Li, and Song]{mi_gan_3}
Yuheng Zhang, Ruoxi Jia, Hengzhi Pei, Wenxiao Wang, Bo~Li, and Dawn Song.
\newblock The secret revealer: Generative model-inversion attacks against deep neural networks.
\newblock In \emph{IEEE / CVF Computer Vision and Pattern Recognition Conference (CVPR)}, 2020.

\bibitem[Zhao et~al.(2021)Zhao, Zhang, Xiao, and Lim]{mi_attack_1}
Xuejun Zhao, Wencan Zhang, Xiaokui Xiao, and Brian~Y. Lim.
\newblock Exploiting explanations for model inversion attacks.
\newblock In \emph{International Conference on Computer Vision (ICCV)}, 2021.

\bibitem[Zhu et~al.(2017)Zhu, Park, Isola, and Efros]{cyclegan}
Jun{-}Yan Zhu, Taesung Park, Phillip Isola, and Alexei~A. Efros.
\newblock Unpaired image-to-image translation using cycle-consistent adversarial networks.
\newblock In \emph{International Conference on Computer Vision (ICCV)}, 2017.

\end{thebibliography}
}

\end{document}